\documentclass{article} 
\usepackage{locowm,times}
\iclrfinalcopy

\usepackage[utf8]{inputenc} \usepackage[T1]{fontenc}    \usepackage[colorlinks]{hyperref}       \usepackage{url}            \usepackage{booktabs}       \usepackage{amsfonts}              \usepackage{microtype}      \usepackage{xcolor}         
\definecolor{nhred}{HTML}{D62728}
\hypersetup{
 colorlinks=True,
 linkcolor=nhred,
 citecolor=gray,
 urlcolor=nhred,
}

\usepackage{graphicx}

\usepackage[font=small]{caption}

\usepackage{bm}
\usepackage{bbm}
\usepackage{array}

\usepackage{float}
\usepackage{placeins}
\usepackage{amsmath}
\usepackage{multirow}

\usepackage{titletoc}

\usepackage{enumitem}
\usepackage{cleveref}
\usepackage{indentfirst}

\newcommand{\tb}[1]{\textbf{#1}}    \newcommand{\tn}[1]{\underline{#1}} 

\graphicspath{{figs/}}

\newcommand{\Aref}[1]{
  {Appendix~\ref{#1}}
}

\title{LocoWM: High-Precision Locomotion through\\ World-Model-Guided Residual Adaptation}

\author{
  \small
    \textbf{Zijie Zhao}$^{1,2,\dagger}$,
    \textbf{Shengqian Chen}$^{2,1,\dagger}$,
    \textbf{Xiaoxu Wang}$^{3}$,
    \textbf{Han Jiang}$^{4}$,
    \textbf{Yuanheng Zhu}$^{2,1,\ast}$,
    \textbf{Dongbin Zhao}$^{2,1}$ \\
    \small $^{1}$University of Chinese Academy of Sciences \\
    \small $^{2}$Institute of Automation, Chinese Academy of Sciences \\
    \small $^{3}$Beijing University of Posts and Telecommunications\\
    \small $^{4}$Beijing Jiaotong University \\
    \small $^{\dagger}$Equal contribution.
}

\begin{document}

\maketitle

\vspace{-20pt}

\begin{figure}[ht]
  \centering
  \includegraphics[width=1.0\textwidth]{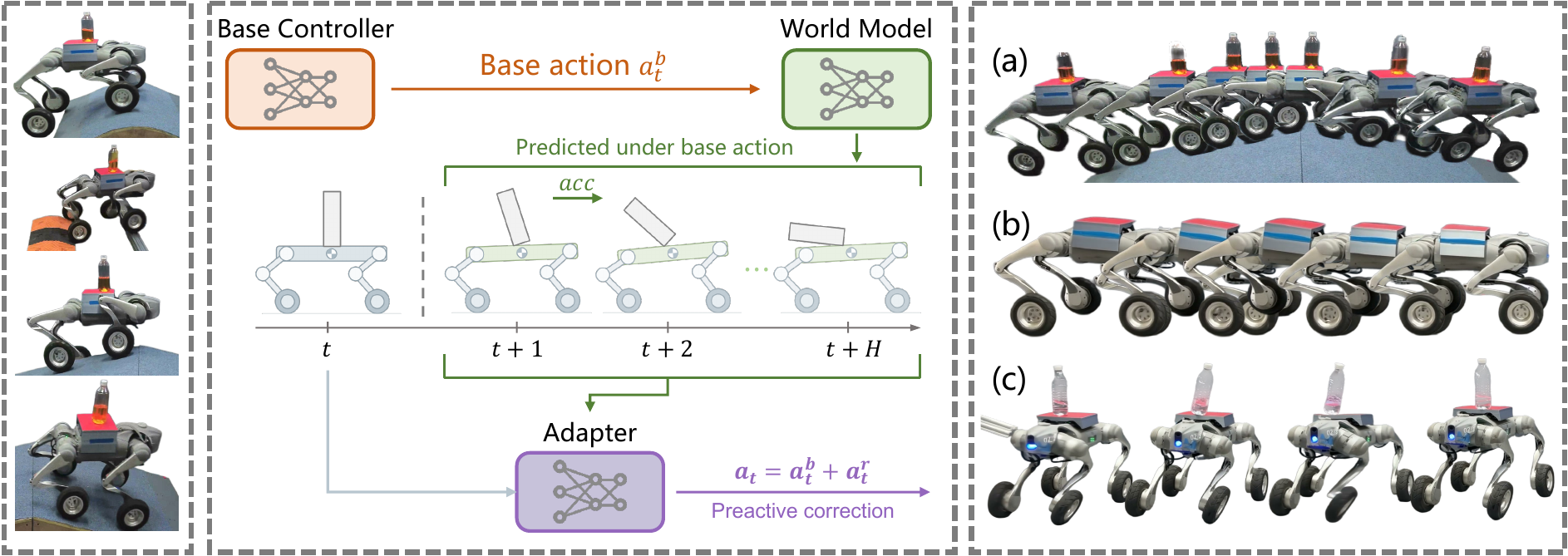}
  \caption{
    \textbf{Left:} Real-world transport of unsecured payloads across varied terrains.
    \textbf{Center:} LocoWM combines the base controller, action-conditioned world model, and residual adapter in a sequential pipeline: the base controller proposes $a_t^b$, the world model predicts future task substates conditioned on $a_t^b$, and the residual adapter uses these predictions to generate a preactive correction $a_t^r$, yielding the executed action $a_t=a_t^b+a_t^r$.
    \textbf{Right:} Real-world demonstrations of (a) terrain leveling, (b) acceleration compensation, and (c) push recovery.
  }
  \label{fig:teaser}
\end{figure}

\vspace{-5pt}
\begin{abstract}
\vspace{-5pt}
High-precision locomotion combines motion-command tracking with precise regulation of task-relevant physical states, enabling robots to interact reliably with their surroundings during motion.
Joint end-to-end optimization can leave precision objectives insufficiently optimized, while reactive residual control adjusts actions only after deviations become observable.
We present \textbf{LocoWM}, a world-model-guided preactive residual adaptation framework for high-precision locomotion.
A base policy provides command-following locomotion, while an action-conditioned world model predicts a sequence of future physical states from proprioceptive history and the proposed base action.
A residual adapter conditions on this predicted sequence to generate additive action corrections that compensate for anticipated deviations.
Two-stage training first learns locomotion and action-conditioned dynamics, then freezes both modules while training the adapter, separating locomotion acquisition from precision adaptation.
Experiments spanning terrain leveling, acceleration compensation, and push recovery demonstrate improved control precision and disturbance robustness over end-to-end and reactive residual baselines.
Demos and code are available at: \href{https://zhaozijie2022.github.io/LocoWM/}{LocoWM}.
\end{abstract}

\vspace{-10pt}

\section{INTRODUCTION}
\label{sec:introduction}
\vspace{-5pt}
Reinforcement learning has enabled agile legged locomotion~\citep{hwangbo2019learning,lee2020learning}, efficient skill acquisition~\citep{rudin2022learning}, and traversal of challenging terrain with legs and wheels~\citep{miki2022learning,lee2024learning}.
Many tasks additionally require precise regulation of task-relevant body poses or physical states while following motion commands.
For example, humanoids may need to stabilize end effectors while walking~\citep{jang2025seec}, and wheeled-legged robots may need to coordinate body orientation with acceleration~\citep{kanno2025tilt}.
These tasks exemplify the challenge of \emph{high-precision locomotion}: meeting task-specific precision requirements while following motion commands.
Joint end-to-end optimization can yield competent locomotion while leaving task-specific precision objectives insufficiently optimized~\citep{huang2026steadytray}.
To address this imbalance, precision learning can be separated from locomotion acquisition, allowing task-specific compensation for locomotion-induced disturbances.

Residual control provides a modular way to improve task precision by learning corrections to a base controller~\citep{yu2022multimodal,zhao2025resmimic}.
Existing observation-conditioned residual controllers generate corrections from current and historical measurements~\citep{jang2025seec,huang2026steadytray}.
Such \emph{reactive} feedback responds to observed deviations without explicitly predicting their subsequent evolution.
Forecasting future task states enables \emph{preactive} correction, allowing the controller to compensate for anticipated deviations before they become observable~\citep{liu2026worldmodelaugmented}.
Model predictive control (MPC) implements this anticipatory principle by using analytical dynamics to predict future states and optimize actions online~\citep{dicarlo2018dynamic}.
Its effectiveness depends on model accuracy and computational resources, while unmodeled dynamics can degrade precision and robustness~\citep{cheng2025rambo,zhou2025adaptive}.

Learning dynamics directly from interaction~\citep{chua2018deep,ha2018world} can supply the forecasts needed for preactive control without relying on an analytical model.
Recent world-model methods learn latent transition dynamics and use them to predict future states for control~\citep{lai2025wmp,li2026lingbotva,zhang2026lingbotva2}.
This learned dynamics interface exposes action-conditioned future consequences in a form that a downstream controller can consume directly.

We introduce LocoWM, a framework for high-precision locomotion through world-model-guided residual adaptation.
The central idea is to forecast the physical quantities defining task precision, termed \emph{task substates}, and use their anticipated evolution to guide action corrections.
As illustrated in the center of \Cref{fig:teaser}, the base policy proposes an action, and the world model predicts a short future substate sequence from this action and proprioceptive history.
The residual adapter combines this forecast with current observations and the base action to generate a correction for anticipated deviations, yielding the executed action $a_t=a_t^b+a_t^r$.

We evaluate LocoWM on a Go2-W wheeled-legged robot through the three tasks illustrated on the right of \Cref{fig:teaser}.
Terrain leveling tests posture regulation under uneven support, acceleration compensation tests the coordination of body tilt with acceleration and braking, and push recovery tests stability recovery after external disturbances.
Our main contributions are:
\begin{itemize}[leftmargin=1.5em, itemsep=1pt, topsep=2pt]
  \item We introduce a physically meaningful task-substate interface for LocoWM that unifies the precision objective, world-model prediction target, and residual-control input around the same task-relevant quantities.
  \item We develop a two-stage preactive residual architecture that first learns locomotion and task-substate dynamics, then freezes both modules while training an adapter for task precision and command following. 
  At inference, the future sequence is predicted in one forward pass and directly guides action corrections without online trajectory optimization.
  \item Experiments demonstrate higher task success rates than end-to-end policies and reactive residual baselines. Relative to the current-state reconstruction baseline, LocoWM improves success by up to $29.6$ percentage points and achieves maximum reductions of $91.8\%$ in roll RMS error and $83.0\%$ in peak vertical acceleration across the evaluated conditions.
\end{itemize}
\vspace{-10pt}

\section{Related Work}
\vspace{-5pt}

\textbf{High-Precision Locomotion and Residual Adaptation.}
High-precision locomotion couples motion-command tracking with the stabilization of a task-relevant body part or payload.
SEEC~\citep{jang2025seec} and Hold My Beer~\citep{li2025hold} stabilize humanoid end effectors under locomotion-induced disturbances, while SteadyTray~\citep{huang2026steadytray} learns to balance unsecured objects during tray transport.
For wheeled-legged robots, base-angle adjustment can compensate for acceleration-induced payload tilt~\citep{kanno2025tilt}.
Residual learning provides a modular way to adapt an existing controller to such task-specific objectives~\citep{yu2022multimodal,ankile2025residual}.
Existing residual methods can be broadly divided according to the information used to condition the correction.
Observation-conditioned residuals derive corrections from current and historical measurements~\citep{jang2025seec,huang2026steadytray}; because they react only after a deviation becomes observable, they are inherently reactive.
To anticipate the effect of an action before the resulting deviation occurs, model-based residual controllers use analytical dynamics models to forecast future states, often within MPC~\citep{jeon2025residual,cheng2025rambo,zhou2025adaptive}.
Analytical-model look-ahead is brittle under model mismatch and unmodeled contact or payload dynamics.
LocoWM instead uses a learned short-horizon dynamics predictor to condition preactive residual corrections on action-conditioned future task substates in a single forward pass.

\textbf{World Models for Embodied AI.}
World models learn dynamics from interaction and support embodied control through several routes. 
\emph{Imagined rollouts} use predicted trajectories to generate synthetic experience for policy learning~\citep{li2025roboticwm,levy2026simdist,wu2022daydreamer}. 
\emph{Online planning} predicts future transitions and optimizes action sequences at decision time~\citep{hansen2024hierarchical,shirwatkar2024piploco,lai2025wmp}. 
\emph{Predictive representation learning} uses future-state modeling objectives to construct features for downstream policies~\citep{zhang2025track,danesh2026morphology,choi2026dawn,yan2026efficiently}. 
LocoWM uses predicted task-relevant future states at inference time to condition a residual controller. 
Specifically, it forecasts action-conditioned task substates in one forward pass and maps them to an additive action correction without online action-sequence optimization.

\vspace{-10pt}
\section{Methodology}
\vspace{-10pt}

We present LocoWM, a framework for high-precision locomotion that combines a locomotion policy, an action-conditioned world model, and a residual adapter.
The world model maps proprioceptive history and the base policy's proposed action to a short sequence of task substates in one forward pass.
The residual adapter uses this sequence to generate additive action corrections without online action optimization.
By conditioning on predicted future task substates, LocoWM shifts residual control from \emph{reactive} correction of observed deviations to \emph{preactive} correction of anticipated ones.

\vspace{-5pt}
\subsection{High-Precision Locomotion} 
\label{subsec:high-precision-locomotion}
\vspace{-5pt}

\textbf{Problem Formulation.}
We consider a \emph{high-precision locomotion} task that simultaneously satisfies two objectives: (1) tracking the user command, and (2) precisely regulating a task-relevant joint or physical quantity throughout locomotion.
Examples include keeping a mounted platform level~\citep{huang2026steadytray}, stabilizing an end effector~\citep{li2025hold,jang2025seec}, and limiting the tilt of an unsecured payload~\citep{kanno2025tilt}.

We model the task as a goal-conditioned Markov decision process $(\mathcal{S}, \mathcal{O}, \mathcal{G}, \mathcal{A}, \mathcal{Z}, P, r, \gamma)$. 
At time $t$, $s_t\in\mathcal{S}$ denotes the full state, including the robot, terrain, and any mounted payload or platform; 
command $g_t\in\mathcal{G}$ specifies the desired robot motion;
action $a_t\in\mathcal{A}$ specifies leg-joint position targets and wheel-joint velocity targets for the low-level controllers;
observation $o_t=[s^{\mathrm{prop}}_{t-L:t},\, g_{t-L:t},\, a_{t-L-1:t-1}]\in\mathcal{O}$ stacks an $L$-step history of proprioceptive signals, user commands, and previous actions, which constitute the information available to the policy at deployment;
transition dynamics $P(s_{t+1}\mid s_t,a_t)$ govern how the full state evolves;
reward $r = r^{\mathrm{track}} + r^{\mathrm{prec}}$ combines a command-tracking term and a precision term,  
and $\gamma\in(0,1)$ is the discount factor.

We introduce a \emph{task substate} $z_t=h(s_t)\in\mathcal{Z}$, where $h$ projects the full state onto a semantically meaningful quantity that must be kept precise; 
$z_t$ may be privileged, computable during training but absent from the observation $o_t$.
It plays three roles and this is what unifies LocoWM: 
(i)~it \textbf{defines the precision objective} through a reward term $r^{\mathrm{prec}}(z_t,z^*_t)$, where $z^*_t$ is the desired precision target; 
(ii)~it is \textbf{what the world model predicts}: LocoWM predicts $\hat z_{:+H}$ as the task-relevant consequence of the current locomotion action;\footnote{~$:+H$ denotes $t+1:t+H$}
and (iii)~its prediction $\hat z_{:+H}$ \textbf{conditions the residual adapter}, enabling preactive correction of predicted deviations.

\textbf{Task Instantiation.}
In this paper, we instantiate the formulation on a legged-wheel robot that carries a tray on its back~\citep{huang2026steadytray}. 
The goal is to regulate the back orientation so that the mounted tray counteracts the robot's motion, staying level during steady locomotion and tilting to absorb inertial forces under acceleration, thereby ensuring steady payload transportation~\citep{kanno2025tilt}.
The task substate is defined as $z_t= \left[\bm{\theta}, \bm{a}, \bm{\omega}\right]$, $\bm{\theta}=[\theta_p, \theta_r]$ denote the back's pitch and roll angles, $\bm{a}$ are the linear accelerations, and $\bm{\omega}=\dot{\bm{\theta}}$ are the angular velocities.
The high-precision objective requires the components of $z_t$ to jointly satisfy
\begin{equation}
  \label{eq:ideal-proj-grav}
  \begin{bmatrix}\theta_p \\ \theta_r \end{bmatrix} = \arctan \left(\frac{-1}{\sqrt{a_{x}^2 + a_{y}^2 + g^2}} \begin{bmatrix} a_{x} \\ a_{y} \end{bmatrix}\right),
  \quad
  [\bm{\omega}, a_z] = \bm{0},
\end{equation}
where $g$ is the gravitational acceleration.
Geometrically, this objective aligns the tray normal with the \emph{effective gravity}, which is the resultant of true gravity and the inertial reaction to acceleration.
Under this objective, the payload experiences no in-plane force and therefore does not slide or tip over, while $[\bm{\omega}, a_z]=\bm{0}$ further ensures steady transportation by eliminating oscillations about the target orientation and vertical jolts.
The detailed task definition and Implementation can be found in \Aref{sec:app-task}.

\subsection{LocoWM: World-Model Guided Residual Adaptation}
\label{subsec:framework}

\begin{figure}[ht]
  \centering
  \includegraphics[width=1.0\textwidth]{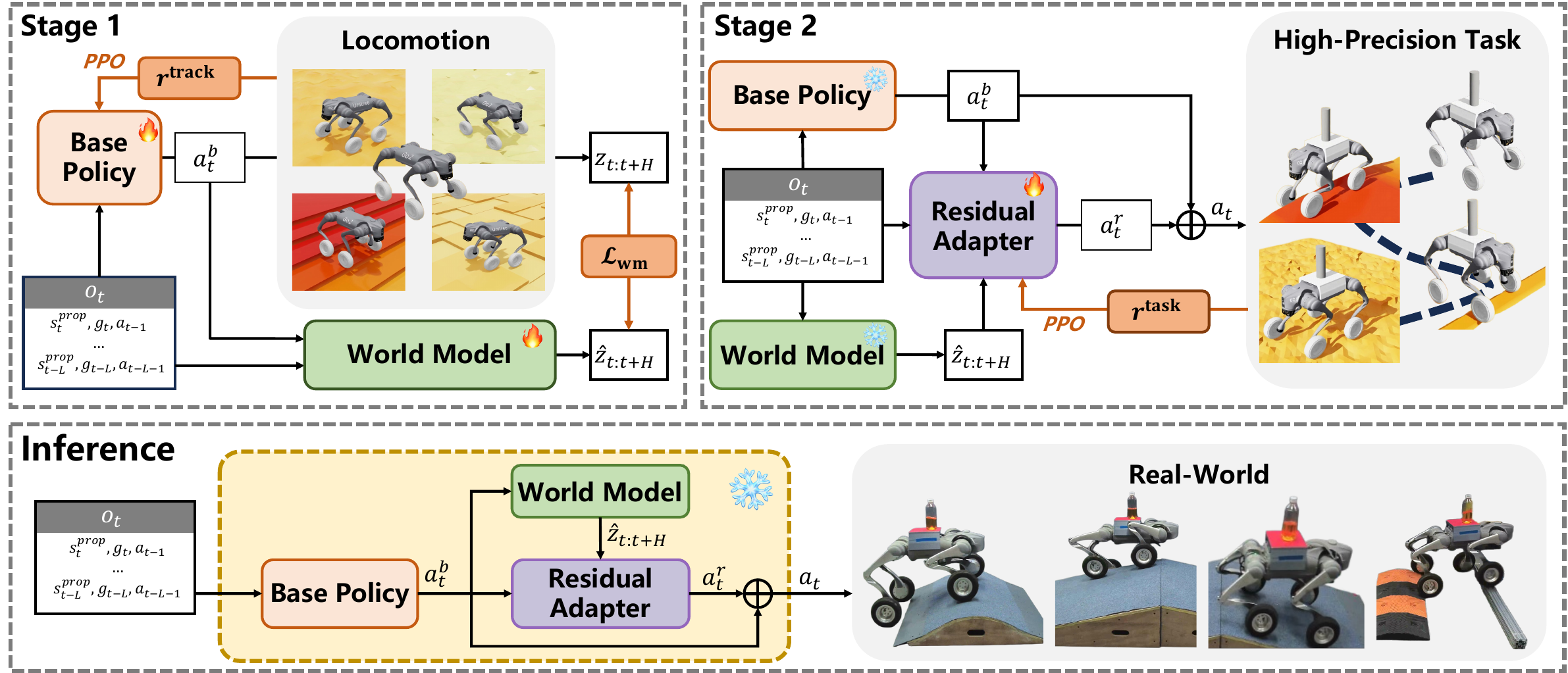}
  \caption{
    Overview of LocoWM.
    \textbf{Stage~1: Base Policy \& World Model.} The base policy learns command-following locomotion, while the world model predicts future task substates from the same trajectories using a separate loss.
    \textbf{Stage~2: Residual Adapter.} The base policy and world model are frozen, and the adapter learns precision corrections.
    \textbf{Inference.} All modules are frozen; the base action, predicted substates, and residual correction are computed sequentially, yielding $a_t=a^b_t+a^r_t$.
  }
  \label{fig:method}
\end{figure}

LocoWM separates locomotion acquisition from precision adaptation through the two training stages in \Cref{fig:method}.
The base policy supplies the motion prior, the world model predicts its task-relevant consequences, and the residual adapter learns corrections on top of both frozen modules.
This factorization is essential for the precision objective: if tracking and precision were optimized jointly from the outset, the dense tracking reward could dominate the sparser precision signal, allowing the robot to keep moving without regulating the tray~\citep{huang2026steadytray}.
We therefore train the base policy with the tracking objective alone, freeze it together with the world model, and then learn a lightweight adapter that cannot perturb the locomotion prior.

\textbf{Stage 1: Learning locomotion and task-substate dynamics.}
The first stage learns the motion prior and the task-substate dynamics from a shared rollout stream.
We train the base policy $\pi^b(o_t)$ with PPO~\citep{schulman2017ppo} using $r^{\mathrm{track}}$, which combines velocity tracking with regularization of motion smoothness, joint and torque limits, and contact behavior.
The regularization encourages smooth, low-agitation motion, which defines the task-substate dynamics that the world model must forecast and the residual adapter must later compensate.
An erratic base policy would make both prediction and precision control harder.
To obtain a reusable prior under the conditions encountered after adaptation, Stage~1 randomizes robot's physical parameters, terrain, and external pushes.

Alongside the base policy, we train an action-conditioned world model on the same rollout data,
\begin{equation}
  \label{eq:stage-1}
  f_\psi:(o_t,a^b_t)\longmapsto\hat z_{:+H},
  \qquad a^b_t=\pi^b(o_t),
\end{equation}
which predicts the next task-substate sequence in a single forward pass.
The prediction target $z_{:+H}$ is collected along the base-policy trajectory and is available only during training.
The world model minimizes
\begin{equation}
  \label{eq:wm-loss}
  \mathcal{L}_{\mathrm{WM}}(\psi)
  =\mathbb{E}_{(o_t,a^b_t,z_{:+H})\sim\pi^b}
  \left[\left\lVert f_\psi(o_t,a^b_t)-z_{:+H}\right\rVert^2\right].
\end{equation}
The base policy and world model share interaction data but have separate optimization objectives.
The prediction gradient updates only $\psi$ and never propagates into the base policy.

Three design properties make $f_\psi$ particularly well suited to guide the residual and distinguish LocoWM from prior world-model designs.
\begin{enumerate}[leftmargin=1.5em, label=\textbullet, itemsep=0pt, topsep=0pt]
\item \textbf{Causal and on-distribution dynamics}: because $f_\psi$ is trained on the base-policy rollout stream and is queried with the same observation--base-action interface at inference, its training inputs match the interface used at deployment.
Its predictions therefore describe how the system will evolve under the base policy’s actions, rather than the behavior of an oracle controller or a generic forward model.
\item \textbf{Observable-to-semantic prediction}: the input $(o_t, a^b_t)$ is purely deployable, while the output is the low-dimensional task substate itself rather than the full state or a latent code. 
Predicting only the task substate keeps $f_\psi$ physically interpretable: $\hat z_{:+H}$ corresponds directly to future tilt and angular velocity, without requiring an analytical dynamics model.
\item \textbf{Single-shot prediction}: the entire chunk $\hat z_{:+H}$ is emitted in one forward pass, so LocoWM can use the forecast to guide the adapter without online rollouts, or action-sequence optimization as in planner-based world models~\citep{hansen2024hierarchical,shirwatkar2024piploco}.
\end{enumerate}

\textbf{Stage 2: Learning precision through residual adaptation.}
After Stage~1, we freeze both $\pi^b$ and $f_\psi$ and train a residual adapter
\begin{equation}
  \pi^r:(o_t,a^b_t,\hat z_{:+H})\longmapsto a^r_t,
  \qquad a_t=a^b_t+a^r_t.
\end{equation}
The adapter's output layer is zero-initialized so that $a^r_t\approx\mathbf{0}$.
At the start of Stage~2, the combined controller therefore has the same mean action as the frozen base policy, preserving the locomotion prior and providing a stable starting point.
Corrective behavior emerges gradually through training instead of perturbing a competent controller from the first update.

Only the adapter and its value function are updated in Stage~2.
PPO uses the combined reward $r^{\mathrm{track}}_t+r^{\mathrm{prec}}_t$, where $r^{\mathrm{prec}}_t$ penalizes deviation of the measured task substate from the high-precision objective.
Because the base policy already follows the command $g_t$, the adapter can focus on regulating the task substate.
The tracking term prevents corrective actions that would compromise command following.
This stage thus optimizes precision on top of a fixed locomotion prior rather than changing the base-policy parameters.
Reward terms, randomization, and curricula are detailed in Appendix~\ref{sec:app-settings}.

\textbf{Inference: Predicting and correcting within the feedback loop.}
At each control step, proprioceptive measurements, commands, and previous actions update the observation history $o_t$.
The frozen modules are evaluated in order:
\begin{equation}
  \label{eq:inference}
  a^b_t=\pi^b(o_t),\qquad
  \hat z_{:+H}=f_\psi(o_t,a^b_t),\qquad
  a^r_t=\pi^r(o_t,a^b_t,\hat z_{:+H}).
\end{equation}
The sum $a_t=a^b_t+a^r_t$ is mapped to motion targets for the low-level controllers.
The next observation closes the feedback loop, while no module is updated during deployment.
The forecast describes task-substate evolution under the base-policy continuation and is provided to the adapter alongside the observation and the proposed base action.
The complete forecast is produced once per step, and the adapter maps it directly to an action correction without iterative model rollouts or online action search.

\textbf{From reactive to preactive control.}
Methods that derive corrections from current observations, observation histories, or a reconstructed current state can act only after a deviation appears in the observations and are therefore reactive~\citep{jang2025seec,huang2026steadytray}.
LocoWM adds the action-conditioned prediction $\hat z_{:+H}$, making future task-substate evolution an explicit input to the correction.
For example, a predicted tilt at time $t+1$ can inform a residual action at time $t$, before that deviation becomes observable.
This changes the residual controller from correcting an observed error to correcting an anticipated consequence of the proposed locomotion action.

\textbf{Physical interface and modularity.}
The three modules are composed in native, physically meaningful spaces rather than through a shared latent representation.
At the input, the predicted substate $\hat z_{:+H}$ lies in the same physical space as the observation $o_t$, allowing the adapter to ground the forecast directly against its sensory input.
At the output, the correction is applied as an additive residual $a_t=a^b_t+a^r_t$ in the base policy's action space rather than through feature-wise (FiLM-style) modulation.
The modules therefore do not require a shared encoder, which decouples their optimization and keeps the base policy, world model, and adapter independently trainable in their respective stages.
The same task substate consequently links the precision objective, the prediction target, and the control input, while each module retains a separate training objective.

\section{Experiments}
\label{sec:experiments}

Our evaluation addresses four questions.
\textbf{(Q1)}~Does LocoWM improve terrain leveling and payload retention across terrains, and how does it affect command tracking (\Cref{subsec:exp-terrain})?
\textbf{(Q2)}~Does LocoWM coordinate platform tilt with acceleration and braking to compensate for inertial forces (\Cref{subsec:exp-tilt-resp})?
\textbf{(Q3)}~How does payload-retention success vary with the direction and magnitude of external pushes (\Cref{subsec:exp-push})?
\textbf{(Q4)}~Do these precision-control behaviors transfer zero-shot to the physical robot (\Cref{subsec:exp-real})?

\subsection{Experimental Setup}
\label{subsec:exp-setup}

Our main experiments run on a Unitree Go2-W wheeled-legged quadruped carrying a rigid, back-mounted stabilization platform 30$\times$20~cm.
\footnote{~To examine whether the same design extends beyond this embodiment, we additionally evaluate the framework on bimanual humanoid tray transport; the experimental setup and results are provided in Appendix~\ref{sec:app-humanoid}.}
At deployment the policy reads proprioception only, with no external motion capture and no platform-mounted force sensor.
\footnote{~Additional implementation details are provided in \Aref{sec:app-settings}.}

\textbf{Terrains.}
For the tasks defined in \Cref{subsec:high-precision-locomotion} and Appendix~\ref{sec:app-task}, we evaluate on four terrain families, each designed to induce a distinct disturbance to the robot's pose.
\begin{enumerate}[leftmargin=1.5em, label=\textbullet, itemsep=-2pt, topsep=-3pt]
  \item \textbf{Slopes}: require the robot to maintain a stable body pitch while climbing and descending.
  \item \textbf{Bumps}: repeatedly perturb the body as the robot traverses uneven ground.
  \item \textbf{One-sided bridge}: creates a persistent height difference between the left and right sides, requiring the robot to maintain a level body.
  \item \textbf{Rough terrain}: introduces continuous irregular disturbances.
\end{enumerate}

\textbf{Baselines.}
We compare LocoWM against four baselines that progressively add its key components. 
This design isolates the contribution of each component: 
\emph{End-to-End $\rightarrow$ React} demonstrates the benefit of residual adaptation;
\emph{React $\rightarrow$ Recon} measures the value of privileged state reconstruction;
\emph{Recon $\rightarrow$ LocoWM} tests the benefit of preactive control by replacing current-state reconstruction with future task-substate prediction.
\begin{enumerate}[leftmargin=1.5em, label=\textbullet, itemsep=-2pt, topsep=-3pt]
\item \textbf{Base Policy}: The frozen locomotion policy trained solely on the velocity tracking reward.
\item \textbf{End-to-End}: A monolithic, single-stage policy trained directly on both tracking and precision rewards.
\item \textbf{React}~\citep{huang2026steadytray}: A reactive residual adapter conditioned exclusively on deployable observations $o_t$.
\item \textbf{Recon}~\citep{sun2025wmr}: Extends \emph{React} with the reconstructed current task substate $\hat z_t$, retaining reactive control without an explicit future-state forecast.
\end{enumerate}

\textbf{Metrics.}
We evaluate platform precision, command tracking, and payload retention using four metrics.
Formal definitions, units, and aggregation conventions are provided in Appendix~\ref{subsec:app-metrics}.
\begin{enumerate}[leftmargin=1.5em, label=\textbullet, itemsep=-2pt, topsep=-3pt]
  \item \textbf{Platform angle error.} Pitch and roll errors measure deviations from the desired orientation.
  RMS and peak errors characterize sustained deviations and transient excursions, respectively.
  \item \textbf{Vertical acceleration.} Mean and peak acceleration magnitudes characterize ride smoothness and vertical impacts that can disturb the payload.
  \item \textbf{Command tracking error.} The discrepancy between commanded and measured velocity captures locomotion performance and its trade-off with payload stabilization.
  \item \textbf{Success rate.} The fraction of trials retaining the payload throughout the evaluation interval measures transport success.
\end{enumerate}

\subsection{Terrain Leveling}
\label{subsec:exp-terrain}
We command the robot to gradually accelerate to a target velocity and then traverse each terrain at approximately constant speed. 
The metrics are reported in \Cref{tab:terrain}.
LocoWM attains the most precise leveling on all terrains.
A slope creates a fore--aft height difference, so climbing and descending load the \emph{pitch} axis.
LocoWM reduces the pitch RMS to 2.61$^\circ$, compared with 28.18$^\circ$ for the Base Policy.
A one-sided bridge raises the wheels on one side and therefore loads the \emph{roll} axis.
In this setting, LocoWM cuts the roll RMS to 0.92$^\circ$, an order of magnitude below the next-best Recon and over 30$\times$ below the Base Policy.
Bumps introduce abrupt contact disturbances, for which peak vertical acceleration measures impact attenuation. 
LocoWM limits $|a_z|_{\max}$ to 3.97~m/s$^2$, compared with 10.17~m/s$^2$ for Recon, indicating improved impact attenuation over current-state reconstruction.
These lower errors relative to the reactive React and Recon baselines support the usefulness of future task-substate predictions for preactive residual correction during terrain traversal.

Higher payload-retention success can coexist with larger command-tracking errors, reflecting the trade-off between command following and payload stabilization.
LocoWM matches the task-agnostic Base Policy closely and stays the lowest of all residual adapters, even overtaking the Base Policy on the bridge.
Together, precise leveling and a smoother ride accompany the highest reported payload retention on every terrain, reaching $98.1\%$ on the bridge and $87.5\%$ on bumps.

\begin{table}[ht]
\caption{Terrain leveling across four terrains. Within each terrain, \textbf{bold} and \underline{underline} mark the best and second best.}
\label{tab:terrain}
\centering
\begin{scriptsize}
\setlength{\tabcolsep}{4pt}
\begin{tabular}{@{}llcccccccc@{}}
\toprule
\multicolumn{2}{l}{Simulation Results} & \multicolumn{2}{c}{Pitch (deg)} & \multicolumn{2}{c}{Roll (deg)} & \multicolumn{2}{c}{$|a_z|$ (m/s$^2$)} & \multirow{2}{*}{$e^{\mathrm{track}}$}  & \multirow{2}{*}{Succ (\%)}  \\
\cmidrule(lr){3-4}\cmidrule(lr){5-6}\cmidrule(lr){7-8}
\textbf{Terrain} & \textbf{Method} & RMS & peak & RMS & peak & mean & peak &  & \\
\midrule
\multirow{5}{*}{Slope}  & Base Policy           & 28.18     & 29.98      & 9.96       & 21.51      & 1.58      & 8.54       & \tb{0.28} & 45.3      \\
                        & End-to-End            & 15.18     & 32.38      & 2.99       & \tn{6.15}  & 0.89      & 7.19       & \tn{0.42} & \tn{62.8} \\
                        & React                 & 5.39      & 20.89      & 2.15       & 8.16       & 0.83      & 7.12       & 0.43      & 59.8      \\
                        & Recon                 & \tn{3.01} & \tn{11.97} & \tn{1.25}  & 7.14       & \tn{0.74} & \tn{5.94}  & 0.46      & 61.5      \\
                        & \textbf{LocoWM(ours)} & \tb{2.61} & \tb{3.90}  & \tb{0.31}  & \tb{1.23}  & \tb{0.41} & \tb{3.10}  & 0.45      & \tb{91.1} \\
\cmidrule(l){1-10}
\multirow{5}{*}{Bump}   & Base Policy           & 1.06      & 14.25      & 2.26       & 14.06      & 3.46      & 19.61      & \tb{0.65} & 21.1      \\
                        & End-to-End            & 0.99      & 17.04      & 1.31       & 15.41      & 3.06      & 19.47      & \tn{0.65} & 50.2      \\
                        & React                 & 0.56      & \tn{10.52} & 0.28       & \tn{10.31} & 2.41      & 10.28      & 1.00      & 45.2      \\
                        & Recon                 & \tn{0.42} & 11.51      & \tn{0.11}  & 10.32      & \tn{2.37} & \tn{10.17} & 0.93      & \tn{67.7} \\
                        & \textbf{LocoWM(ours)} & \tb{0.18} & \tb{3.00}  & \tb{0.09}  & \tb{2.47}  & \tb{1.22} & \tb{3.97}  & 0.87      & \tb{87.5} \\
\cmidrule(l){1-10}
\multirow{5}{*}{Bridge} & Base Policy           & 5.91      & 16.90      & 30.81      & 44.40      & 1.91      & 21.27      & \tn{0.43} & 9.1       \\
                        & End-to-End            & 4.81      & 25.56      & 25.66      & 42.73      & \tn{1.31} & \tn{8.66}  & 0.80      & 2.2       \\
                        & React                 & 2.35      & 10.44      & 21.07      & 21.31      & 1.38      & 15.68      & 0.68      & 78.4      \\
                        & Recon                 & \tn{1.56} & \tn{9.36}  & \tn{11.24} & \tn{20.00} & 1.36      & 14.38      & 0.74      & \tn{95.5} \\
                        & \textbf{LocoWM(ours)} & \tb{0.91} & \tb{2.52}  & \tb{0.92}  & \tb{3.54}  & \tb{0.27} & \tb{2.45}  & \tb{0.33} & \tb{98.1} \\
\cmidrule(l){1-10}
\multirow{5}{*}{Rough}  & Base Policy           & 4.47      & 12.47      & 7.46       & 10.47      & 1.43      & 8.55       & \tb{0.13} & 91.1      \\
                        & End-to-End            & 3.01      & 10.19      & 2.12       & 14.93      & 1.01      & 6.74       & 0.42      & 95.4      \\
                        & React                 & \tn{1.84} & 9.63       & 2.24       & 5.86       & 1.00      & \tb{6.07}  & 0.33      & 95.6      \\
                        & Recon                 & 2.08      & \tn{9.04}  & \tn{1.34}  & \tn{4.12}  & \tn{0.98} & 6.37       & 0.25      & \tn{96.2} \\
                        & \textbf{LocoWM(ours)} & \tb{1.73} & \tb{4.35}  & \tb{1.19}  & \tb{4.07}  & \tb{0.93} & \tn{6.17}  & \tn{0.20} & \tb{97.8} \\

\bottomrule
\end{tabular}\end{scriptsize}
\vspace{-10pt}
\end{table}

\subsection{Acceleration Compensation}
\label{subsec:exp-tilt-resp}

\begin{figure}[ht]
\centering
\includegraphics[width=0.8\textwidth]{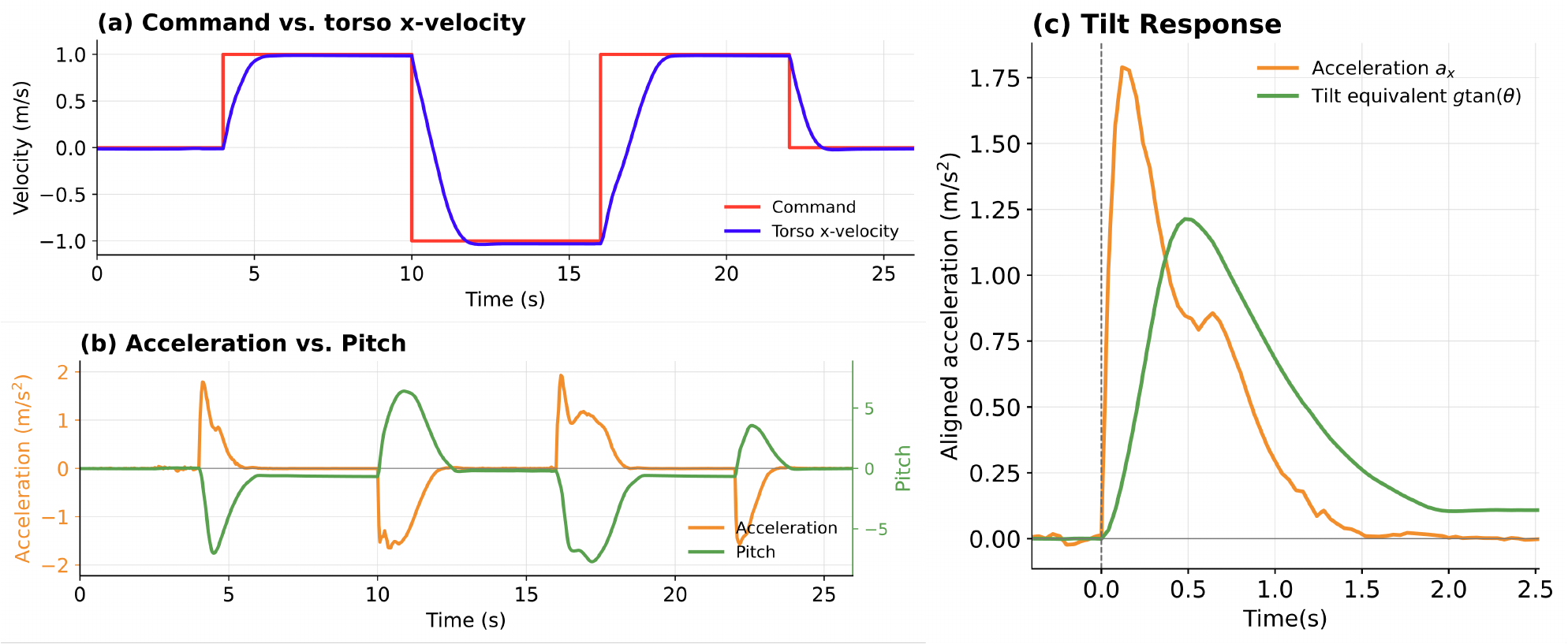}
\caption{
  Acceleration compensation on flat ground. 
  \textbf{(a)}~Torso $x$-velocity tracking; 
  \textbf{(b)}~Acceleration--pitch responses; 
  \textbf{(c)}~Single-event tilt response.
}
\label{fig:tilt-accel}
\end{figure}

We additionally evaluate the acceleration compensation task, in which the robot repeatedly accelerates and brakes on flat ground. 
Unlike the terrain leveling task, this maneuver requires the body to actively pitch to counteract the inertial forces generated by acceleration and braking.
We command the robot to track a $\pm1$~m/s step-velocity profile, and as shown in \Cref{fig:tilt-accel}(a), the torso velocity follows both the positive and negative velocity commands. 
Over the resulting repeated acceleration and deceleration events, \Cref{fig:tilt-accel}(b) shows that each change in acceleration is accompanied by a corresponding pitch response. 
The single-event trace in \Cref{fig:tilt-accel}(c) further shows that the tilt-equivalent acceleration $g\tan\theta$ follows the torso acceleration $a_x$ with the appropriate sign, indicating that LocoWM banks into the acceleration. 
This coordinated response is consistent with LocoWM's preactive control: the residual adapter uses the world model's prediction of upcoming in-plane acceleration $(a_x,a_y)$ to guide pitch correction, producing the compensation condition \(g\tan\theta^\star=a\) in \Cref{eq:ideal-proj-grav}. 
The platform therefore remains aligned with the effective gravity direction, allowing a freely placed payload to remain stable without sliding or toppling.

\subsection{Push Recovery}
\label{subsec:exp-push}

\begin{figure}[ht]
\centering
\includegraphics[width=0.9\textwidth]{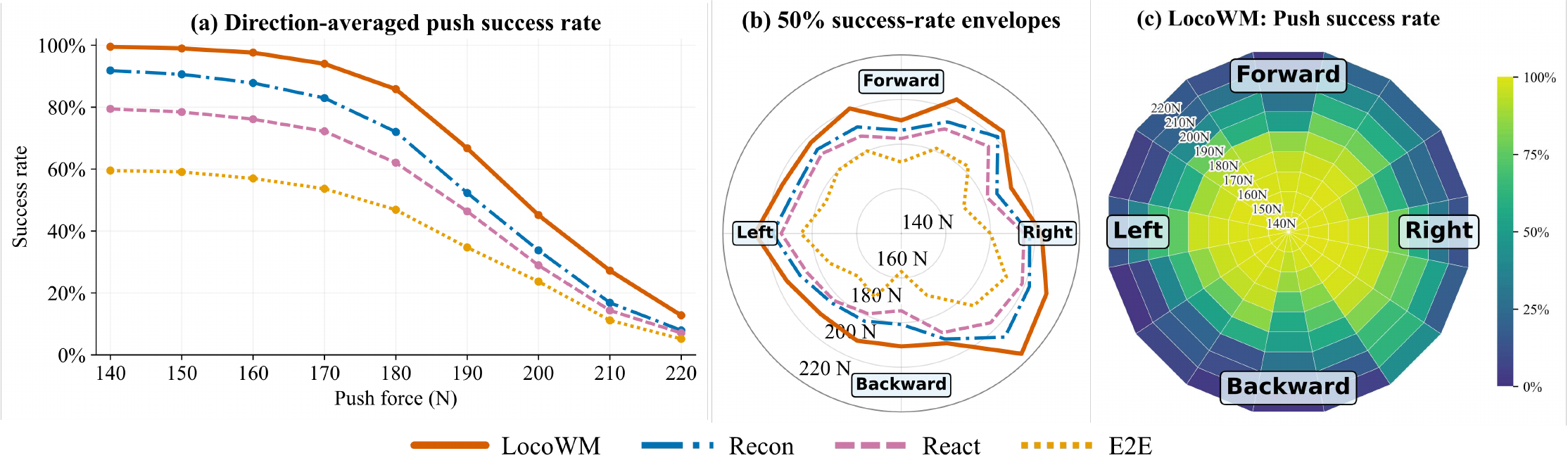}
\caption{
  Payload retention during push recovery.
  \textbf{(a)}~Direction-averaged success rates versus push magnitude.
  \textbf{(b)}~The $50\%$ success-rate envelopes of all methods.
  \textbf{(c)}~LocoWM success-rate map: angle denotes push direction, radius denotes force magnitude, and color denotes success rate.
}
\label{fig:push2}
\end{figure}

In the push recovery task, we drive the robot forward at 0.5~m/s while applying pushes to its torso, varying both push direction and magnitude and recording the payload-retention success rate.
As shown in \Cref{fig:push2}(a), the direction-averaged success rate decreases as the push force increases for all methods, whereas LocoWM consistently achieves the highest success rate across the tested force range.\Cref{fig:push2}(b) further reveals the directional dependence of this robustness: all methods tolerate larger lateral pushes than forward or backward pushes, and LocoWM maintains the outermost $50\%$ success-rate envelope in nearly every direction. 
The corresponding success-rate map for LocoWM in \Cref{fig:push2}(c) exhibits the same anisotropic pattern, with high success rates extending to larger force magnitudes along the lateral directions than along the forward and backward directions. 
Together, these results show that LocoWM's preactive residual controller provides a broader payload-retention range under the evaluated disturbances. 
The success-rate maps of the baseline methods are provided in \Aref{subsec:app-sim-result}.

\subsection{Real-World Results}
\label{subsec:exp-real}

We deploy LocoWM zero-shot on the physical robot and demonstrate terrain leveling, acceleration compensation, and push recovery.
\footnote{~The remaining real-world experimental results are available in \Aref{subsec:app-real}. }

\begin{figure}[ht]
\centering
    \centering
    \includegraphics[width=1.0\linewidth]{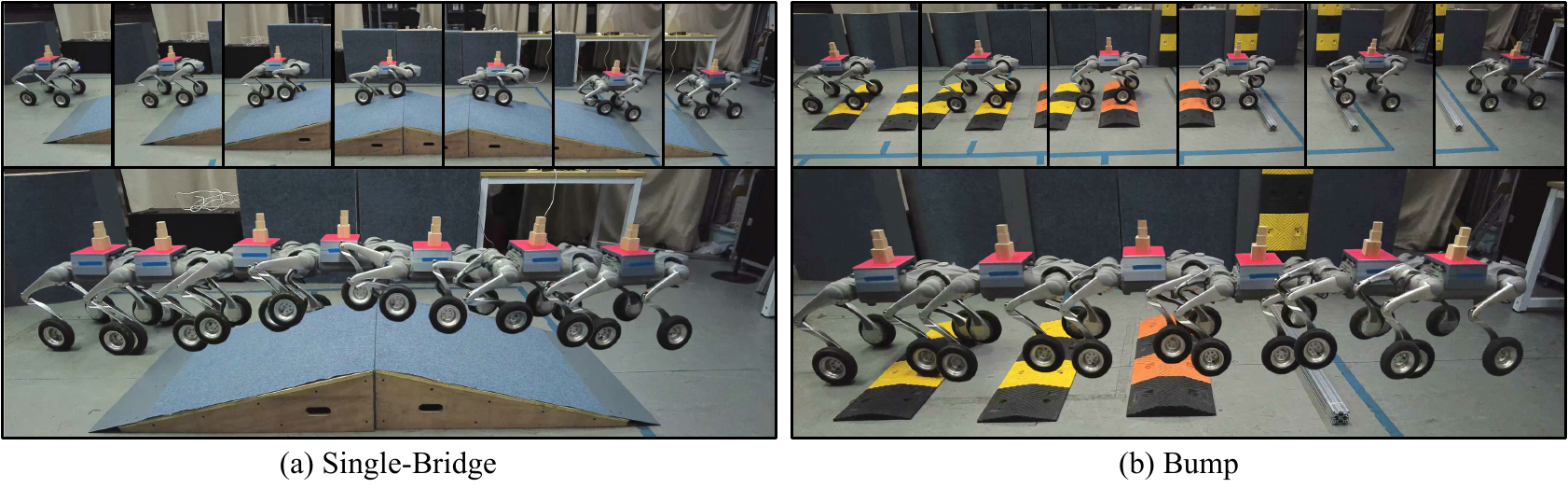}
    \caption{
        Real-world terrain leveling on (a) a one-sided bridge and (b) bumps.
        In each panel, the top row shows successive video frames, while the bottom row combines robot cutouts extracted from individual frames onto a common background to illustrate the motion sequence.
    }
\label{fig:real-leveling}
\end{figure}
\vspace{-5pt}

In the terrain leveling task, the robot carries an unsecured payload across a one-sided bridge, slopes, bumps, and wavy terrain.
These surfaces combine gradual height changes with abrupt contacts, inducing sustained body inclination and coupled roll--pitch disturbances.
As shown in \Cref{fig:real-leveling}, LocoWM reconfigures the body and legs as the support height changes, keeping the back-mounted platform approximately level in both representative sequences while the payload remains upright.

\begin{figure}[ht]
\centering
    \centering
    \includegraphics[width=1.0\linewidth]{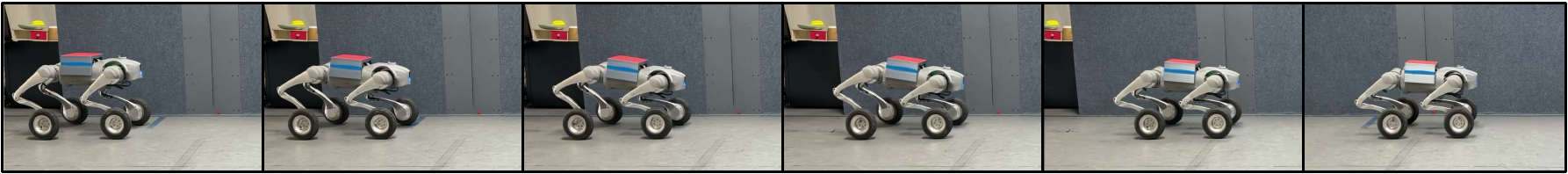}
    \caption{
        Real-world acceleration compensation on flat ground.
    }
\label{fig:real-tilt-resp}
\end{figure}

In the acceleration compensation task, the robot accelerates and decelerates on flat ground. 
The sequence in \Cref{fig:real-tilt-resp} shows the platform pitching with the longitudinal motion, providing a real-world counterpart to the coordinated tilt response observed in simulation. 
The body therefore adjusts its orientation as the motion changes rather than leaving the platform fixed during acceleration.

\begin{figure}[ht]
\centering
    \centering
    \includegraphics[width=1.0\linewidth]{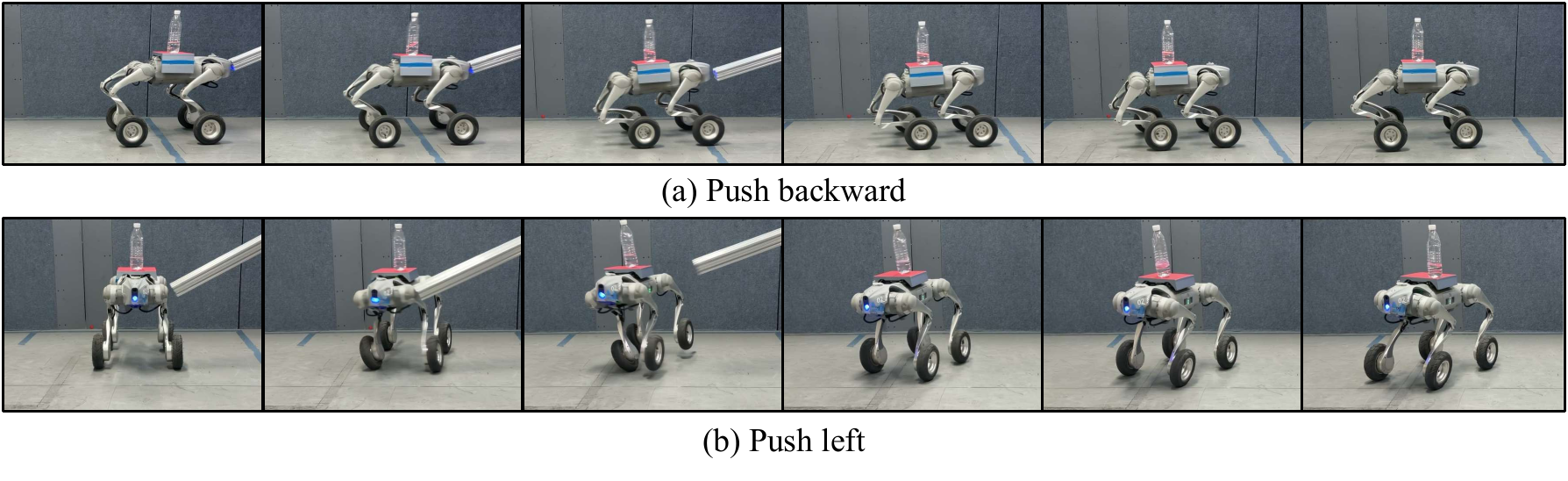}
    \caption{
        Real-world push recovery with an unsecured bottle.
        Frames progress from left to right through force application, the transient response, and recovery; the bottle remains on the platform in both sequences.
    }
\label{fig:real-push}
\end{figure}
In the push recovery task, we apply backward and leftward forces to the torso while an unsecured bottle rests on the back-mounted platform. 
LocoWM changes the body posture during each disturbance and then brings the platform back toward level. 
Although the bottle tilts transiently, it remains on the platform throughout both recovery sequences. 
Together, these demonstrations show that the preactive residual controller transfers its precision behaviors to hardware without task-specific retraining.

\section{Conclusion}

We presented LocoWM, a world-model-guided preactive residual adaptation framework for high-precision locomotion.
The framework combines a base locomotion policy, an action-conditioned world model that forecasts future task substates, and a residual adapter that uses these forecasts to refine the base action.
Experiments on Go2-W demonstrate higher task success rates than end-to-end and reactive residual baselines.
Relative to current-state reconstruction, success improves by up to $29.6$ percentage points across the evaluated terrains, alongside lower posture errors and reduced vertical impacts.
Acceleration compensation and push recovery further demonstrate coordinated body tilting and stability recovery after external disturbances.

\textbf{Limitations.}
The main experiments center on the Go2-W wheeled-legged platform and three predefined task settings.
To examine the generality of LocoWM beyond this setting, \Aref{sec:app-humanoid} further provides a simulation-based analysis on bimanual Unitree G1 humanoid transport.

\paragraph{Future Work.}
Future work will further investigate future physical-state prediction as an interface between established motor skills and task-specific precision objectives.
This direction will examine how prediction-guided adaptation can extend existing motion skills to broader task settings while preserving command following and precision control.
It will also explore richer task-substate representations and longer-horizon prediction for capturing more complex physical consequences.

\clearpage
\section*{Ethics Statement}
This work studies high-precision robot locomotion and stable payload transport through learned dynamics and residual control.
No human participants, animal subjects, or personal data were involved.
The physical experiments with the Go2-W platform were conducted in a controlled environment with appropriate safety precautions.

\section*{Reproducibility Statement}
The LocoWM architecture and two-stage training procedure are described in \Cref{subsec:framework}.
Appendix~\ref{sec:app-task} specifies the task substates and precision objectives.
Appendix~\ref{sec:app-settings} details the simulation and control settings, observations, prediction targets, rewards, domain randomization, and curricula.
The evaluation tasks, baselines, and metrics are defined in \Cref{subsec:exp-setup}, with additional experimental results provided in Appendix~\ref{sec:app-additional-results}.

\section*{Use of AI}
AI was used for writing assistance and language polishing during manuscript preparation.
The authors reviewed and verified all AI-assisted content and take full responsibility for the research claims, experimental results, and final content of this paper.

{
  \footnotesize
  \bibliography{locowm}

@inproceedings{li2025hold,
  title = {{Hold My Beer: Learning Gentle Humanoid Locomotion and End-Effector Stabilization Control}},
  author = {Li, Yitang and Zhang, Yuanhang and Xiao, Wenli and Pan, Chaoyi and Weng, Haoyang and He, Guanqi and He, Tairan and Shi, Guanya},
  booktitle = {Proceedings of The 9th Conference on Robot Learning},
  year = {2025},
  volume = {305},
  pages = {4506--4523},
  series = {Proceedings of Machine Learning Research},
  publisher = {PMLR},
  url = {https://proceedings.mlr.press/v305/li25i.html}
}

@article{schulman2017ppo,
  title = {{Proximal Policy Optimization Algorithms}},
  author = {Schulman, John and Wolski, Filip and Dhariwal, Prafulla and Radford, Alec and Klimov, Oleg},
  journal = {arXiv preprint arXiv:1707.06347},
  year = {2017},
  eprint = {1707.06347},
  archivePrefix = {arXiv},
  url = {https://arxiv.org/abs/1707.06347}
}

@article{zhao2025resmimic,
  title = {{ResMimic: From General Motion Tracking to Humanoid Whole-body Loco-Manipulation via Residual Learning}},
  author = {Zhao, Siheng and Ze, Yanjie and Wang, Yue and Liu, C. Karen and Abbeel, Pieter and Shi, Guanya and Duan, Rocky},
  journal = {arXiv preprint arXiv:2510.05070},
  year = {2025},
  eprint = {2510.05070},
  archivePrefix = {arXiv},
  url = {https://arxiv.org/abs/2510.05070}
}

@inproceedings{zhang2025track,
  title = {{Track Any Motions under Any Disturbances}},
  author = {Zhang, Zhikai and Guo, Jun and Chen, Chao and Wang, Jilong and Lin, Chenghuai and Lian, Yunrui and Xue, Han and Wang, Zhenrong and Liu, Maoqi and Lyu, Jiangran and Liu, Huaping and Wang, He and Yi, Li},
  booktitle = {2026 IEEE International Conference on Robotics and Automation (ICRA)},
  year = {2026},
  organization = {IEEE},
  url = {https://arxiv.org/abs/2509.13833}
}

@article{lee2024learning,
  title = {{Learning robust autonomous navigation and locomotion for wheeled-legged robots}},
  author = {Lee, Joonho and Bjelonic, Marko and Reske, Alexander and Wellhausen, Lorenz and Miki, Takahiro and Hutter, Marco},
  journal = {Science Robotics},
  year = {2024},
  volume = {9},
  number = {89},
  pages = {eadi9641},
  url = {https://doi.org/10.1126/scirobotics.adi9641}
}

@article{huang2026steadytray,
  title = {{SteadyTray: Learning Object Balancing Tasks in Humanoid Tray Transport via Residual Reinforcement Learning}},
  author = {Huang, Anlun and Wu, Zhenyu and Atar, Soofiyan and Zhi, Yuheng and Yip, Michael},
  journal = {arXiv preprint arXiv:2603.10306},
  year = {2026},
  eprint = {2603.10306},
  archivePrefix = {arXiv},
  url = {https://arxiv.org/abs/2603.10306}
}

@inproceedings{jang2025seec,
  title = {{SEEC: Stable End-Effector Control with Model-Enhanced Residual Learning for Humanoid Loco-Manipulation}},
  author = {Jang, Jaehwi and Wang, Zhuoheng and Zhou, Ziyi and Wu, Feiyang and Zhao, Ye},
  booktitle = {2026 IEEE International Conference on Robotics and Automation (ICRA)},
  year = {2026},
  organization = {IEEE},
  url = {https://arxiv.org/abs/2509.21231}
}

@inproceedings{ankile2025residual,
  title = {{Residual Off-Policy RL for Finetuning Behavior Cloning Policies}},
  author = {Ankile, Lars and Jiang, Zhenyu and Duan, Rocky and Shi, Guanya and Abbeel, Pieter and Nagabandi, Anusha},
  booktitle = {2026 IEEE International Conference on Robotics and Automation (ICRA)},
  year = {2026},
  organization = {IEEE},
  url = {https://arxiv.org/abs/2509.19301}
}

@article{jeon2025residual,
  title = {{Residual MPC: Blending Reinforcement Learning With GPU-Parallelized Model Predictive Control}},
  author = {Jeon, Se Hwan and Lee, Ho Jae and Hong, Seungwoo and Kim, Sangbae},
  journal = {IEEE Transactions on Robotics},
  year = {2026},
  volume = {42},
  pages = {3469--3488},
  url = {https://doi.org/10.1109/tro.2026.3721669}
}

@article{cheng2025rambo,
  title = {{RAMBO: RL-Augmented Model-Based Whole-Body Control for Loco-Manipulation}},
  author = {Cheng, Jin and Kang, Dongho and Fadini, Gabriele and Shi, Guanya and Coros, Stelian},
  journal = {IEEE Robotics and Automation Letters},
  year = {2025},
  volume = {10},
  number = {9},
  pages = {9462--9469},
  url = {https://doi.org/10.1109/lra.2025.3594984}
}

@article{zhou2025adaptive,
  title = {{Adaptive Legged Locomotion via Online Learning for Model Predictive Control}},
  author = {Zhou, Hongyu and Zhang, Xiaoyu and Tzoumas, Vasileios},
  journal = {IEEE Robotics and Automation Letters},
  year = {2026},
  volume = {11},
  number = {2},
  pages = {1778--1785},
  url = {https://doi.org/10.1109/lra.2025.3644161}
}

@inproceedings{li2026lingbotva,
  title = {{Causal World Modeling for Robot Control}},
  author = {Li, Lin and Zhang, Qihang and Luo, Yiming and Yang, Shuai and Wang, Ruilin and Zhang, Luyao and Yu, Mingrui and Gao, Zelin and Xue, Nan and Zhou, Boyu and Zhu, Xing and Ding, Mingyu and Shen, Yujun and Xu, Yinghao},
  booktitle = {Proceedings of Robotics: Science and Systems},
  year = {2026},
  address = {Sydney, Australia},
  month = {jul},
  url = {https://www.roboticsproceedings.org/rss22/p016.html}
}

@article{zhang2026lingbotva2,
  title = {{Native Video-Action Pretraining for Generalizable Robot Control}},
  author = {Zhang, Qihang and Li, Lin and Zhang, Luyao and Yang, Shuai and Luo, Yiming and Li, Shuaiting and Wang, Ruilin and Wang, Junke and Shao, Jiahao and Xu, Gangwei and Zhou, Jiaming and Shen, Yishu and Jin, Yudong and Xu, Fangyi and Ma, Shuailei and Liao, Jiaqi and Lu, Guanxing and Shi, Zifan and Wen, Yongkun and Zhao, Yujie and Tang, Weixuan and Wang, Xinyang and Li, Chaojian and Zhu, Jiapeng and Cheng, Ka Leong and Xue, Nan and Zhu, Xing and Shen, Yujun and Xu, Yinghao},
  journal = {arXiv preprint arXiv:2607.08639},
  year = {2026},
  eprint = {2607.08639},
  archivePrefix = {arXiv},
  url = {https://arxiv.org/abs/2607.08639}
}

@inproceedings{lai2025wmp,
  title = {{World Model-Based Perception for Visual Legged Locomotion}},
  author = {Lai, Hang and Cao, Jiahang and Xu, Jiafeng and Wu, Hongtao and Lin, Yunfeng and Kong, Tao and Yu, Yong and Zhang, Weinan},
  booktitle = {2025 IEEE International Conference on Robotics and Automation (ICRA)},
  year = {2025},
  pages = {11531--11537},
  url = {https://doi.org/10.1109/icra55743.2025.11128762}
}

@article{sun2025wmr,
  title = {{Learning Humanoid Locomotion with World Model Reconstruction}},
  author = {Sun, Wandong and Chen, Long and Su, Yongbo and Cao, Baoshi and Liu, Yang and Xie, Zongwu},
  journal = {arXiv preprint arXiv:2502.16230},
  year = {2025},
  eprint = {2502.16230},
  archivePrefix = {arXiv},
  url = {https://arxiv.org/abs/2502.16230}
}

@article{li2025roboticwm,
  title = {{Robotic World Model: A Neural Network Simulator for Robust Policy Optimization in Robotics}},
  author = {Li, Chenhao and Krause, Andreas and Hutter, Marco},
  journal = {arXiv preprint arXiv:2501.10100},
  year = {2025},
  eprint = {2501.10100},
  archivePrefix = {arXiv},
  url = {https://arxiv.org/abs/2501.10100}
}

@inproceedings{levy2026simdist,
  title = {{Simulation Distillation: Pretraining World Models in Simulation for Rapid Real-World Adaptation}},
  author = {Levy, Jacob and Westenbroek, Tyler and Huang, Kevin and Palafox, Fernando and Yin, Patrick and Omidshafiei, Shayegan and Kim, Dong-Ki and Gupta, Abhishek and Fridovich-Keil, David},
  booktitle = {Proceedings of Robotics: Science and Systems},
  year = {2026},
  address = {Sydney, Australia},
  month = {jul},
  url = {https://www.roboticsproceedings.org/rss22/p017.html}
}

@inproceedings{hansen2024hierarchical,
  title = {{Hierarchical World Models as Visual Whole-Body Humanoid Controllers}},
  author = {Hansen, Nicklas and {Jyothir S V} and Sobal, Vlad and LeCun, Yann and Wang, Xiaolong and Su, Hao},
  booktitle = {International Conference on Learning Representations},
  year = {2025},
  url = {https://iclr.cc/virtual/2025/poster/30793}
}

@inproceedings{shirwatkar2024piploco,
  title = {{PIP-Loco: A Proprioceptive Infinite Horizon Planning Framework for Quadrupedal Robot Locomotion}},
  author = {Shirwatkar, Aditya and Saxena, Naman and Chandra, Kishore and Kolathaya, Shishir},
  booktitle = {2025 IEEE International Conference on Robotics and Automation (ICRA)},
  year = {2025},
  pages = {11198--11204},
  url = {https://doi.org/10.1109/icra55743.2025.11128382}
}

@inproceedings{kanno2025tilt,
  title = {{Relative Tilt Suppression of a Carried Object Using Base Link Angle Adjustment on a Quadruped-Wheeled Robot}},
  author = {Kanno, Kimikage and Hashimoto, Kenji and Mizuuchi, Ikuo},
  booktitle = {2025 IEEE/RSJ International Conference on Intelligent Robots and Systems (IROS)},
  year = {2025},
  pages = {10684--10689},
  url = {https://doi.org/10.1109/iros60139.2025.11247400}
}

@article{hwangbo2019learning,
  title = {{Learning agile and dynamic motor skills for legged robots}},
  author = {Hwangbo, Jemin and Lee, Joonho and Dosovitskiy, Alexey and Bellicoso, Dario and Tsounis, Vassilios and Koltun, Vladlen and Hutter, Marco},
  journal = {Science Robotics},
  year = {2019},
  volume = {4},
  number = {26},
  pages = {eaau5872},
  url = {https://doi.org/10.1126/scirobotics.aau5872}
}

@article{lee2020learning,
  title = {{Learning quadrupedal locomotion over challenging terrain}},
  author = {Lee, Joonho and Hwangbo, Jemin and Wellhausen, Lorenz and Koltun, Vladlen and Hutter, Marco},
  journal = {Science Robotics},
  year = {2020},
  volume = {5},
  number = {47},
  pages = {eabc5986},
  url = {https://doi.org/10.1126/scirobotics.abc5986}
}

@inproceedings{rudin2022learning,
  title = {{Learning to Walk in Minutes Using Massively Parallel Deep Reinforcement Learning}},
  author = {Nikita Rudin and David Hoeller and Philipp Reist and Marco Hutter},
  booktitle = {Proceedings of the 5th Conference on Robot Learning},
  year = {2022},
  volume = {164},
  pages = {91--100},
  series = {Proceedings of Machine Learning Research},
  publisher = {PMLR},
  url = {https://proceedings.mlr.press/v164/rudin22a.html}
}

@article{miki2022learning,
  title = {{Learning robust perceptive locomotion for quadrupedal robots in the wild}},
  author = {Miki, Takahiro and Lee, Joonho and Hwangbo, Jemin and Wellhausen, Lorenz and Koltun, Vladlen and Hutter, Marco},
  journal = {Science Robotics},
  year = {2022},
  volume = {7},
  number = {62},
  pages = {eabk2822},
  url = {https://doi.org/10.1126/scirobotics.abk2822}
}

@inproceedings{dicarlo2018dynamic,
  title = {{Dynamic Locomotion in the {MIT Cheetah 3} Through Convex Model-Predictive Control}},
  author = {Di Carlo, Jared and Wensing, Patrick M. and Katz, Benjamin and Bledt, Gerardo and Kim, Sangbae},
  booktitle = {2018 IEEE/RSJ International Conference on Intelligent Robots and Systems (IROS)},
  year = {2018},
  pages = {1--9},
  url = {https://dspace.mit.edu/handle/1721.1/138000}
}

@inproceedings{ha2018world,
  title = {{Recurrent World Models Facilitate Policy Evolution}},
  author = {Ha, David and Schmidhuber, J{\"u}rgen},
  booktitle = {Advances in Neural Information Processing Systems},
  year = {2018},
  volume = {31},
  publisher = {Curran Associates, Inc.},
  url = {https://dl.acm.org/doi/10.5555/3327144.3327171}
}

@inproceedings{chua2018deep,
  title = {{Deep Reinforcement Learning in a Handful of Trials using Probabilistic Dynamics Models}},
  author = {Chua, Kurtland and Calandra, Roberto and McAllister, Rowan and Levine, Sergey},
  booktitle = {Advances in Neural Information Processing Systems},
  year = {2018},
  volume = {31},
  publisher = {Curran Associates, Inc.},
  url = {https://dl.acm.org/doi/10.5555/3327345.3327385}
}

@misc{liu2026worldmodelaugmented,
  title = {World-Model-Augmented Visual Locomotion for Humanoids on Foothold-Constrained Terrain}, 
  author = {Yuxi Liu and Lijun Han and Ziming Wang and Ao Zhang and Cong Yang and Wei Sui},
  journal = {arXiv preprint arXiv:2609.02542},
  year={2026},
  eprint={2609.02542},
  archivePrefix = {arXiv},
  url = {https://arxiv.org/abs/2609.02542}, 
}

@article{danesh2026morphology,
  title = {Morphology-Conditioned World Model for Cross-Embodiment Quadrupedal Locomotion}, 
  author = {Mohamad H. Danesh and Chenhao Li and Amin Abyaneh and Anas Houssaini and Kirsty Ellis and Glen Berseth and Marco Hutter and Hsiu-Chin Lin},
  journal = {arXiv preprint arXiv:2604.08780},
  year={2026},
  eprint = {2604.08780},
  archivePrefix = {arXiv},
  url = {https://arxiv.org/abs/2604.08780}, 
}

@article{choi2026dawn,
  title = {{DAWN: Noise-Robust Quadruped Parkour via Depth-Denoising World Models}},
  author = {Yohan Choi and Min-Jun Kim and Jin-Sung Kim and Yong-Jae Kim and Youn-Hee Han},
  journal = {arXiv preprint arXiv:2609.29092},
  year = {2026},
  eprint = {2609.29092},
  archivePrefix = {arXiv},
  url = {https://arxiv.org/abs/2609.29092}
}

@article{yan2026efficiently,
  author = {Yan, Yashuai and Egle, Tobias and Ott, Christian and Lee, Dongheui},
  journal = {IEEE Robotics and Automation Letters}, 
  title = {Efficiently Learning Robust Torque-Based Locomotion Through Reinforcement With Model-Based Supervision}, 
  year = {2026},
  volume = {11},
  number = {4},
  pages = {4155-4162},
  url = {https://doi.org/10.1109/lra.2026.3664534}
}

@article{yu2022multimodal,
  title = {{Multi-Modal Legged Locomotion Framework with Automated Residual Reinforcement Learning}},
  journal={IEEE Robotics and Automation Letters}, 
  author = {Yu, Chen and Rosendo, Andre},
  year = {2022},
  volume = {7},
  number = {4},
  pages = {10312-10319},
  url = {https://doi.org/10.1109/lra.2022.3191071}
}

@inproceedings{wu2022daydreamer,
  title = {{DayDreamer: World Models for Physical Robot Learning}},
  author = {Wu, Philipp and Escontrela, Alejandro and Hafner, Danijar and Abbeel, Pieter and Goldberg, Ken},
  booktitle = {Proceedings of the 6th Conference on Robot Learning},
  year = {2022},
  volume = {205},
  pages = {2226--2240},
  series = {Proceedings of Machine Learning Research},
  publisher = {PMLR},
  url = {https://proceedings.mlr.press/v205/wu23c.html}
}
  \bibliographystyle{locowm}
}
\clearpage
\newpage

\begin{center}
   {\LARGE\bfseries APPENDICES}
\end{center}
\startcontents[appendices]
\printcontents[appendices]{l}{0}{\setcounter{tocdepth}{2}}
\newpage

\appendix

\section{Task Definition and Implementation}
\label{sec:app-task}

\subsection{Task Definition.}
\label{subsec:app-task-definition}
\begin{figure}[ht]
\centering
    \centering
    \includegraphics[width=0.95\linewidth]{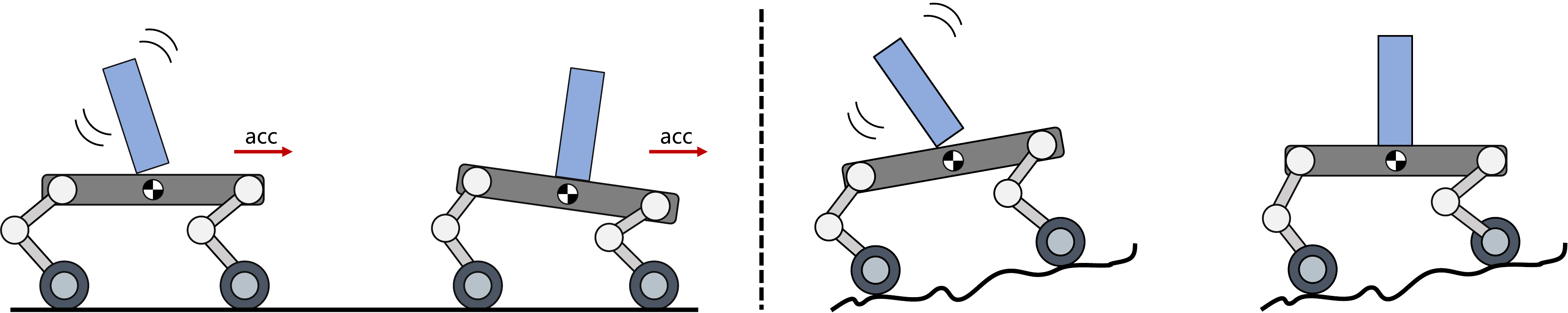}
    \caption{
        Back-mounted platform stabilization.
        \textbf{Left: Acceleration Compensation.}
        Acceleration can destabilize a payload on a level platform; coordinated body tilting compensates for inertial loading.
        \textbf{Right: Terrain Leveling.}
        Uneven support heights disturb the platform orientation; adjustments to the leg configuration counteract torso tilt and maintain a level platform.
    }
\label{fig:app-task-define}
\end{figure}

As described in \Cref{subsec:high-precision-locomotion}, we study the stabilization of a back-mounted platform during wheeled-legged locomotion, with the goal of maintaining the stability of an unsecured payload. 
The platform is rigidly attached to the torso, so its orientation is regulated through coordinated leg and wheel motion.
During acceleration and braking on flat ground, the platform must tilt to compensate for inertial loading and reduce the effective force parallel to its surface (\Cref{fig:app-task-define}, left).
On uneven terrain, changes in wheel contact heights disturb the torso orientation, requiring compensatory motion to keep the platform level (\Cref{fig:app-task-define}, right).
Together, these conditions require the robot to regulate platform orientation according to its motion while tracking locomotion commands.

\subsection{Task Substate Implementation.}
\label{subsec:app-task-substate}
The task involves platform pitch and roll, linear acceleration, and angular velocity.
The implemented task substate uses the nine-dimensional encoding, concatenating base angular velocity, world-frame linear acceleration, and projected gravity:
\[
z_t^{\mathrm{impl}}=\bigl(\boldsymbol\omega_t^{\mathcal B},\boldsymbol a_t^{\mathcal B},\boldsymbol g_t^{\mathrm{proj},\mathcal B}\bigr)\in\mathbb R^9,
\]
where $\mathcal W$ and $\mathcal B$ denote the world and base frames, respectively.
The implemented task substate preserves all the platform tilt, linear acceleration, and tilt-rate information in the definition in \Cref{subsec:high-precision-locomotion}.
Using the ZYX Euler-angle convention, let $R=R_z(\theta_y)R_y(\theta_p)R_x(\theta_r)$ map the base frame to the world frame.
With normalized projected gravity 
\[
\boldsymbol u
=
\frac{\boldsymbol g^{\mathrm{proj},\mathcal B}}
{\|\boldsymbol g^{\mathrm{proj},\mathcal B}\|}
=
R^\top(0,0,-1)^\top,
\] 
the conversion is,
\begin{equation}
\label{eq:task-substate-conversion}
\begin{aligned}
\theta_p&=\operatorname{atan2}\!\left(u_x,\sqrt{u_y^2+u_z^2}\right), \\
\theta_r&=\operatorname{atan2}(-u_y,-u_z),\\
\begin{bmatrix}\dot\theta_p\\\dot\theta_r\end{bmatrix}
&=
\begin{bmatrix}
0 & \cos\theta_r & -\sin\theta_r\\
1 & \sin\theta_r\tan\theta_p & \cos\theta_r\tan\theta_p
\end{bmatrix}
\boldsymbol\omega^{\mathcal B}, \\
\boldsymbol a^{\mathcal W}
&=R\boldsymbol a^{\mathcal B}.
\end{aligned}
\end{equation}

For $|\theta_p|<\pi/2$, these relations recover $z_t=[\boldsymbol\theta_t,\boldsymbol a_t,\dot{\boldsymbol\theta}_t]$ in the main formulation, establishing equivalence for the task quantities.

\subsection{High-Precision Objective.}
\label{subsec:high-precision-objective}
We use the task substate from the main formulation, $z_t=[\bm\theta,\bm a,\bm\omega]$, where $\bm\theta=[\theta_p,\theta_r]$, $\bm a=[a_x,a_y,a_z]$, and $\bm\omega=\dot{\bm\theta}$.
To construct the normalized high-precision target in \Cref{eq:ideal-proj-grav}, consider a translational approximation at the platform reference point, with the world vertical axis pointing upward.
Neglecting local inertial terms induced by platform rotation, gravity and the inertial reaction produce the effective load per unit mass
\[
\bm f_{\mathrm{eff}}=-g\bm e_z-\bm a
=\begin{bmatrix}-a_x&-a_y&-(g+a_z)\end{bmatrix}^{\top}.
\]
For fixed horizontal acceleration, the steady precision target eliminates vertical acceleration and orientation rates, giving $a_z^*=0$ and $\bm\omega^*=\bm0$.
At this target, the effective-load magnitude is $D=\sqrt{a_x^2+a_y^2+g^2}$, and its normalized horizontal components are $[-a_x/D,-a_y/D]^\top$.
The componentwise target mapping in the main formulation sets the tangents of pitch and roll to these normalized components:
\[
\begin{bmatrix}\tan\theta_p^*\\\tan\theta_r^*\end{bmatrix}
=-\frac{1}{D}\begin{bmatrix}a_x\\a_y\end{bmatrix}.
\]
Taking the componentwise inverse tangent and combining the steady-state conditions yields exactly the high-precision target in \Cref{eq:ideal-proj-grav}:
\[
\begin{bmatrix}\theta_p^*\\\theta_r^*\end{bmatrix}
=\arctan\!\left(-\frac{1}{\sqrt{a_x^2+a_y^2+g^2}}
\begin{bmatrix}a_x\\a_y\end{bmatrix}\right),
\qquad [\bm\omega^*,a_z^*]=\bm0.
\]
When $a_x=a_y=0$, the target reduces to a level platform; nonzero horizontal acceleration produces a bounded tilt target through the normalized mapping.

\section{Implementation, Training, and Evaluation Details}
\label{sec:app-settings}
All policies are trained in Isaac~Lab at a $5$~ms physics step ($200$~Hz) and a control decimation of $4$, giving a $50$~Hz control loop; each episode lasts $20$~s. 
The robot is a Unitree Go2-W ($12$ leg joints, $4$ wheels) with the stabilization platform rigidly mounted on the base. 
Leg joints use position control ($K_p{=}50$, $K_d{=}1$) and wheels velocity control ($K_p{=}0$, $K_d{=}0.5$); 
both share a $23.5$~N\,m effort limit, a $30$~rad/s velocity limit. 

\subsection{Observations}
\label{subsec:app-obs}

We use an asymmetric actor--critic. 
The policy observes only deployable proprioception with an $L$-step history ($L{=}6$), $s^{prop}_t=[\,v_{\mathrm c},\omega,g,q{-}q_0,\dot q]_t$;
the critic additionally observes privileged quantities. 
The world model predicts the task substate, emitting the chunk $\hat z_{t+1:t+H}$ in one forward pass. 
Additive uniform noise is applied to the policy group only.

\begin{table}[htbp]
\centering
\caption{Observation terms. Group: P = policy, C = critic, W = world-model target.}
\label{tab:app-obs}
\small
\begin{tabular}{@{}llccccc@{}}
\toprule
Term & Symbol & Group & Dim & Hist.\ & Noise & Scale \\
\midrule
Velocity command & $v_{\mathrm c}$ & P/C & 3 & 6 & -- & 1.0 \\
Base angular velocity & $\omega$ & P/C/W & 3 & 6 & $\pm0.2$ & 0.25 \\
Projected gravity & $g$ & P/C/W & 3 & 6 & $\pm0.05$ & 1.0 \\
Joint positions (legs) & $q{-}q_0$ & P/C & 12 & 6 & $\pm0.01$ & 1.0 \\
Joint velocities & $\dot q$ & P/C & 16 & 6 & $\pm1.5$ & 0.05 \\
Previous action & $a_{t-1}$ & P/C & 16 & 1 & -- & 1.0 \\
Base linear velocity & $v$ & C & 3 & 1 & -- & 2.0 \\
Terrain height scan & -- & C & ${\approx}187$ & 1 & -- & 1.0 \\
Base linear accel.\ (Stage 2) & $a^{\mathrm{base}}$ & C/W & 3 & 6 & -- & 0.25 \\
Ideal projected gravity\ (Stage 2) & $g^{\star}$ & C & 3 & 6 & -- & 1.0 \\
\bottomrule
\end{tabular}
\end{table}

\subsection{World Model Prediction}
\label{subsec:app-world-model-prediction}
Each prediction sample consists of the policy observation, the current base-policy action, and up to $H$ consecutive future substates from the same environment. 
For a sample at rollout index $t$, the label at prediction offset $k \in \{0,\ldots,H-1\}$ is valid only when $t+k<T$ and $\prod_{j=0}^{k}(1-d_{t+j})=1$, where $T$ is the rollout length and $d_t$ indicates episode termination or truncation. 
This mask excludes post-reset states from the future labels of pre-reset samples. 
Samples near an episode boundary retain their valid prediction prefix, with the remaining offsets excluded from the loss.
Future sequences are constructed within each rollout batch without stitching across batches; offsets beyond the batch boundary are zero-filled and masked out. 
Observation histories persist across rollout batches and are reset at episode boundaries. 
When the available history is insufficient, missing entries are filled by repeating the first observation of the new episode.

The acceleration signal used to construct the desired orientation is processed by a first-order exponential IIR low-pass filter, $y_t=\alpha x_t+(1-\alpha)y_{t-1}$, with $\alpha=1-\exp(-2\pi f_c/f_s)$.
The configured cutoff frequency is $f_c=1$~Hz and the control frequency is $f_s=50$~Hz.
Filtering precedes a componentwise hard dead band: horizontal acceleration components with magnitude below $0.1$~m/s$^2$ are set to zero.
The resulting components are multiplied by $1.5$ before constructing the desired gravity direction.
The filter is initialized from the first raw sample.
Reset-marked calls reinitialize the filter only when they bypass its same-step cache; under the current reward-before-reset evaluation order, automatic episode resets retain the cached filter state.

\subsection{Rewards}
\label{subsec:app-reward}

\Cref{tab:app-reward} lists the reward terms for both stages ($\sigma^2{=}0.25$). 
Stage~2 drops the keep-level terms (flat orientation, base height) in favor of a projected-gravity tracking reward, so the platform banks into accelerations rather than staying flat. 
The ideal projected gravity $g^{\star}$ is constructed from the low-pass-filtered ($2$~Hz) horizontal base acceleration with a $0.1$~m/s$^2$ dead-band.

\begin{table}[htbp]
\centering
\caption{Reward terms across the two training stages. A single value is shared by both stages.}
\label{tab:app-reward}
\small
\begin{tabular}{@{}llr@{}}
\toprule
Component & Expression & Weight (S1\,/\,S2) \\
\midrule
Lin.\ vel.\ tracking ($x$) & $\exp(-(v_x^{\mathrm{cmd}}-v_x)^2/\sigma^2)$ & 0.75\,/\,1.0 \\
Lin.\ vel.\ tracking ($y$) & $\exp(-(v_y^{\mathrm{cmd}}-v_y)^2/\sigma^2)$ & 0.75 \\
Ang.\ vel.\ tracking ($z$) & $\exp(-(\omega_z^{\mathrm{cmd}}-\omega_z)^2/\sigma^2)$ & 0.75 \\
Base angle tracking & $\exp(-\lVert g_{xy}-g^{\star}_{xy}\rVert^2/\sigma^2)$ & --\,/\,0.5 \\
Vertical velocity & $v_z^2$ & -2.0\,/\,-10.0 \\
Roll/pitch rate & $\omega_x^2+\omega_y^2$ & -0.05 \\
Flat orientation & $g_x^2+g_y^2$ & -0.5\,/\,-- \\
Joint torques & $\textstyle\sum_j\tau_j^2$ & -2.0e-4 \\
Leg joint accel.\ & $\textstyle\sum_{\mathrm{leg}}\ddot q^2$ & -2.5e-7 \\
Wheel joint accel.\ & $\textstyle\sum_{\mathrm{wheel}}\ddot q^2$ & -2.5e-9 \\
Base height & $(h^{\mathrm base}-0.40)^2$ & -10.0 \\
Undesired contacts & $\sum \mathbbm{1}(\text{non-foot contacts})$ & -1.0 \\
Action rate & $\textstyle\sum(a_t-a_{t-1})^2$ & -0.01 \\
Stand-still (no cmd) & $\mathbbm{1}[\lVert v_{\mathrm c}\rVert{<}0.1]\textstyle\sum_{\mathrm{leg}}|q-q_0|$ & -0.25 \\
Hip deviation & $\textstyle\sum_{\mathrm{hip}}(q-q_0)^2$ & -0.3 \\
Joint deviation (legs) & $\textstyle\sum_{\mathrm{leg}}(q-q_0)^2$ & -0.1 \\
\bottomrule
\end{tabular}
\end{table}

\subsection{Domain Randomization}
\label{subsec:app-domain-randomization}

\Cref{tab:app-dr} lists all randomization. 
Operations are \emph{add} (offset), \emph{scale} (multiplicative), or \emph{abs} (absolute), applied at startup (once per environment), reset (each episode), or on a recurring interval. 
The delayed-PD model lags the applied command by $0$--$3$ physics steps ($0$--$15$~ms).

\begin{table}[htbp]
\centering
\caption{Domain randomization and perturbations. $^{\ast}$log-uniform.}
\label{tab:app-dr}
\small
\begin{tabular}{@{}lrrllc@{}}
\toprule
Term & Min & Max & Unit & Op.\ & When \\
\midrule
Base mass & $-1.0$ & $2.0$ & kg & add & startup \\
Body inertia & $0.5$ & $1.5$ & $\times$ & scale & startup \\
Base CoM offset ($x,y,z$) & $-0.05$ & $0.05$ & m & add & startup \\
Foot static friction & $0.5$ & $1.0$ & -- & abs & startup \\
Foot dynamic friction & $0.5$ & $0.8$ & -- & abs & startup \\
Foot restitution & $0.0$ & $0.5$ & -- & abs & startup \\
Actuator stiffness $K_p$ & $0.8$ & $1.2$ & $\times$ & scale$^{\ast}$ & reset \\
Actuator damping $K_d$ & $0.8$ & $1.2$ & $\times$ & scale$^{\ast}$ & reset \\
Initial joint positions & $0.95$ & $1.05$ & $\times$ & scale & reset \\
Initial base yaw & $-\pi$ & $\pi$ & rad & abs & reset \\
Initial base height & $0.0$ & $0.2$ & m & add & reset \\
Initial base lin.\ vel.\ ($x$) & $-0.5$ & $0.5$ & m/s & abs & reset \\
Initial base lin.\ vel.\ ($y$) & $-0.15$ & $0.15$ & m/s & abs & reset \\
Initial base lin.\ vel.\ ($z$) & $-0.2$ & $0.2$ & m/s & abs & reset \\
Init.\ base ang.\ vel.\ (r/p/y) & $-0.35$ & $0.35$ & rad/s & abs & reset \\
Reset impulse force / torque & $-10$ & $10$ & N,\,N\,m & abs & reset \\
Push velocity ($x$) & $-0.5$ & $0.5$ & m/s & add & interval \\
Push velocity ($y$) & $-0.3$ & $0.3$ & m/s & add & interval \\
\bottomrule
\end{tabular}
\end{table}

\subsection{Curriculum Learning}
\label{subsec:app-curriculum}

\textbf{Terrain Curriculum} follows the standard velocity-based progression: an environment is promoted when it walks past half the terrain length and demoted when it covers less than half its commanded distance.

\textbf{Command Curriculum} is per-environment and per-axis. 
Each axis $i\in\{x,y,z\}$ holds a level $\ell_i\in[0.1,1.0]$ that scales the full command range to $[\ell_i v_i^{\min},\ell_i v_i^{\max}]$. 
Each environment maintains an exponential moving average (EMA) of its per-episode tracking-reward rate,
\[
  \bar r_i\leftarrow(1-\alpha)\bar r_i+\alpha\left(\sum r_i/T_{\mathrm{ep}}\right),
\]
where $T_{\mathrm{ep}}$ is the episode length and $\alpha=0.5$.
At every reset, the command range $\ell_i$ is adjusted by $\pm\Delta$ ($\Delta=0.1$) according to this EMA.
The range is expanded if $\bar r_i>\tau_{\uparrow}w_i$ and shrunk if $\bar r_i<\tau_{\downarrow}w_i$, where $w_i$ is the tracking weight and $(\tau_{\uparrow},\tau_{\downarrow})=(0.8,0.5)$ for the linear axes and $(0.6,0.3)$ for yaw.
As a result, each environment adapts its difficulty independently, and only an environment that consistently tracks commands within its current range is assigned a wider one.

\textbf{Command Resampling.}
At each command resampling ($6$--$10$~s; the command is held at $0$ for the first $50$ steps of every episode), environments are divided into three regimes.
With probability $10\%$, the robot receives a \textbf{zero command} for standing.
With probability $5\%$, it receives a \textbf{bang--bang} step command sampled from the corners of the command box (each axis at min/$0$/max, excluding the all-zero command) to encourage recovery from saturated transients.
The remaining $85\%$ of environments use \textbf{bucket sampling}.

When the command range of an axis has just expanded ($\ell_i>\ell_i^{\mathrm{prev}}$), samples are drawn from three bins with probabilities $(0.15,0.70,0.15)$: the newly opened negative frontier, the previously mastered core, and the newly opened positive frontier.
Otherwise, commands are sampled uniformly from the current range.
This strategy allocates $30\%$ of samples to the newly expanded boundaries while keeping $70\%$ within the previously mastered region, allowing the curriculum to expand progressively without causing a sharp drop in tracking performance.

\subsection{Network Architecture and Optimization}
\label{subsec:app-networks}

\textbf{Network architecture.}
The base policy and world model are MLPs with hidden widths $(512,256,128)$, while the residual adapter uses $(256,128)$.
All three networks use ELU hidden activations and linear output layers.
Each training stage uses a separate critic with hidden widths $(512,256,128)$ and the privileged observations listed in \Cref{tab:app-obs}.
The adapter's output layer is initialized to zero.

\textbf{Optimization.}
In Stage~1, PPO trains the base policy and critic, while a separate Adam optimizer trains the world model on the same rollouts using the prediction loss.
Prediction gradients do not update the base policy.
In Stage~2, the base policy and world model are frozen; PPO updates the residual adapter and a newly initialized critic.
Both stages use the PPO settings in \Cref{tab:app-optimization}.

\begin{table}[htbp]
\centering
\caption{PPO hyperparameters used in both training stages.}
\label{tab:app-optimization}
\small
\begin{tabular}{@{}lc@{}}
\toprule
Hyperparameter & Value \\
\midrule
Optimizer & Adam \\
Initial learning rate & $10^{-3}$ \\
Learning-rate schedule & Adaptive, target KL $0.01$ \\
Rollout length per environment & 24 \\
Epochs per update & 5 \\
Discount factor & 0.99 \\
GAE parameter & 0.95 \\
Clipping parameter & 0.2 \\
Entropy coefficient & 0.01 \\
Value-loss coefficient & 1.0 \\
Clipped value loss & Enabled \\
Maximum gradient norm & 1.0 \\
Initial action standard deviation & 1.0 \\
\bottomrule
\end{tabular}
\end{table}

\textbf{Training and deployment.}
Training uses Isaac~Lab with a physics step of 5~ms, four physics steps per policy update, and a maximum episode duration of 20~s.
The deployment controller is configured to run at 50~Hz, corresponding to a nominal prediction horizon of 100~ms.
All network parameters remain fixed during deployment.

\subsection{Metric Definitions and Aggregation}
\label{subsec:app-metrics}
For each trial $i$, let $\mathcal{T}_i$ denote the valid
measurement samples and $T_i=|\mathcal{T}_i|$.
For a scalar signal $x_{i,t}$, define its root mean square
(RMS), peak magnitude, and mean absolute magnitude as
\begin{equation}
  \label{eq:metric-definitions}
    R_i(x)
      = \sqrt{\frac{1}{T_i}
         \sum_{t\in\mathcal{T}_i}x_{i,t}^{2}}, \qquad
    P_i(x)
      = \max_{t\in\mathcal{T}_i}|x_{i,t}|, \qquad
    A_i(x)
      = \frac{1}{T_i}
         \sum_{t\in\mathcal{T}_i}|x_{i,t}|.
\end{equation}

\textbf{Platform angle error.}
For pitch or roll, indexed by $q\in\{p,r\}$, the angular error is
\begin{equation}
  e_{q,i,t}
  = \operatorname{wrap}_{[-\pi,\pi)}
    \left(\theta_{q,i,t}-\theta^\star_{q,i,t}\right),
\end{equation}
where $\theta^\star_{q,i,t}$ is the desired orientation.
The terrain leveling task uses a horizontal target, $\theta^\star_{p,i,t}=\theta^\star_{r,i,t}=0$.
For the acceleration compensation task, the target depends on acceleration, as specified in \Cref{eq:ideal-proj-grav} and Appendix~\ref{sec:app-task}.

\textbf{Vertical acceleration.}
Let $a_{z,i,t}$ denote translational acceleration along the world vertical axis at the evaluated platform or base reference point.
Its trial-level mean and peak magnitudes are
\begin{equation}
  \bar a_{z,i}=A_i(a_z),
  \qquad
  a^{\mathrm{peak}}_{z,i}=P_i(a_z),
\end{equation}
both expressed in $\mathrm{m/s^{2}}$.

\textbf{Command tracking error.}
For forward-speed tracking, the discrepancy between measured and commanded velocities is
\begin{equation}
  e^{\textrm{track}}
    = R_i \big( v_{i,t}-v^{\mathrm{cmd}}_{i,t} \big),
\end{equation}
where both velocities are expressed in the same coordinate frame.

\textbf{Cross-trial aggregation.}
Let $\mathcal{V}=\{i:T_i>0\}$ denote trials with valid measurement samples.
We compute each continuous-signal metric separately for each trial and report its mean across trials:
\begin{equation}
  \bar m
    = \frac{1}{|\mathcal{V}|}
      \sum_{i\in\mathcal{V}}m_i,
\end{equation}
where $m_i$ is the corresponding trial-level metric.

In particular, peak metrics are obtained by first taking the maximum magnitude within each trial and then averaging these trial-level peaks:
\begin{equation}
  \overline{P}(x)
    = \frac{1}{|\mathcal{V}|}
      \sum_{i\in\mathcal{V}}
      \max_{t\in\mathcal{T}_i}|x_{i,t}|.
\end{equation}
Columns labeled ``max'' in \Cref{tab:terrain} report this mean of per-trial peak magnitudes.
The same aggregation applies to the pitch and roll peak errors and the vertical-acceleration peaks.

\section{Additional Experimental Results}
\label{sec:app-additional-results}

\subsection{Simulation Results}
\label{subsec:app-sim-result}

\FloatBarrier
\subsubsection{Terrain Leveling}
\label{subsubsec:app-terrain-leveling}

\begin{figure}[htbp]
\centering
    \includegraphics[width=0.75\linewidth]{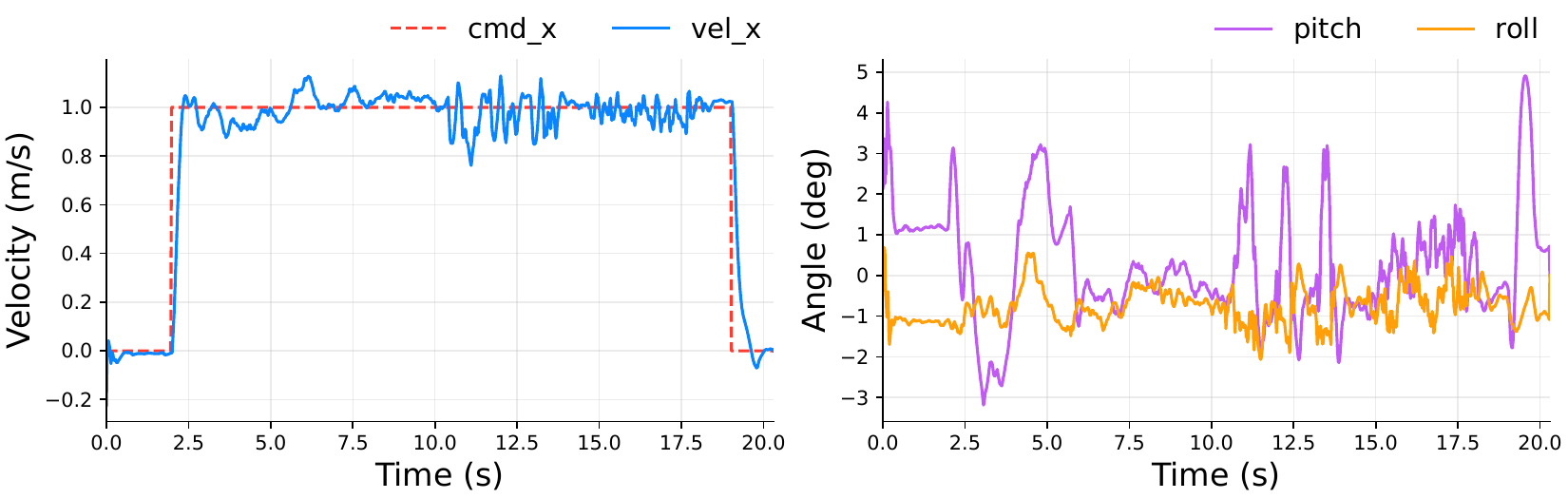}
    \caption{Forward velocity tracking and pitch and roll responses during simulated terrain leveling.}
\label{fig:app-sim-terrain-curve}
\end{figure}

\Cref{fig:app-sim-terrain-curve} shows velocity tracking and orientation responses during terrain leveling with LocoWM.
The forward velocity command increases from $0$ to $1$~m/s at approximately $2$~s and returns to $0$ at approximately $19$~s.
The measured velocity fluctuates around the target during traversal and returns close to zero after the stop command.
The corresponding pitch and roll traces show orientation variations within a few degrees, with larger excursions in pitch than in roll.

\vspace{-5pt}
\vspace{-5pt}

\begin{figure}[htbp]
\centering
    \includegraphics[width=0.85\linewidth]{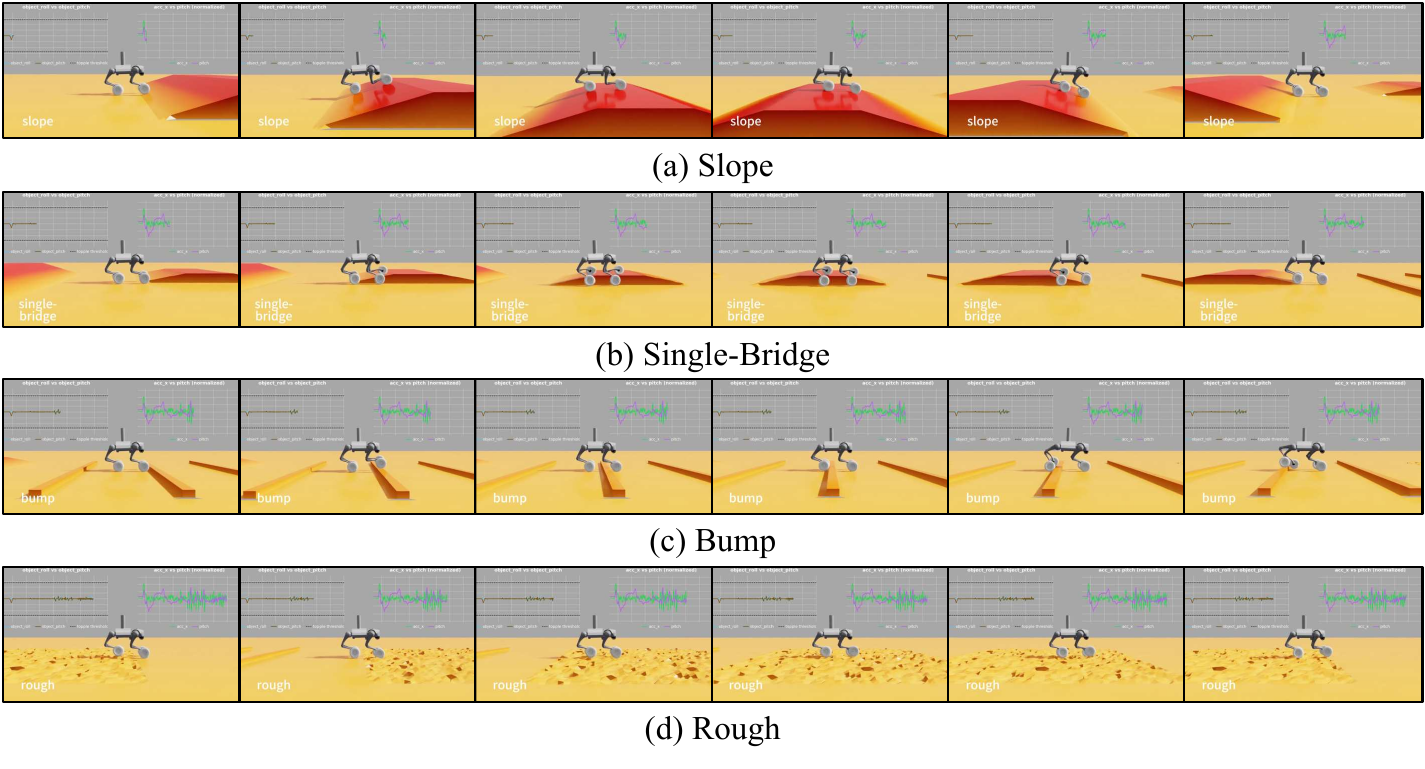}
    \caption{Simulated terrain leveling with an upright payload.}
\label{fig:app-sim-terrain}
\end{figure}

\Cref{fig:app-sim-terrain} presents successive snapshots of payload transport under different support conditions.
On slopes, the front and rear wheels pass through the slope entry, crest, and descent in sequence, while the leg configuration adjusts to changing support heights and the platform remains approximately level.
The one-sided bridge creates a lateral difference in support height, requiring an asymmetric leg configuration to regulate roll.
In both scenes, the payload remains upright throughout the displayed traversal.

Bumps and rough terrain illustrate responses to repeated changes in contact conditions.
Crossing a bump produces distinct support transitions as the front and rear wheels pass over the obstacle, whereas rough terrain introduces continuing irregular disturbances.
The inset torso acceleration and pitch traces fluctuate during these maneuvers, while payload roll and pitch remain within the indicated toppling thresholds.
These sequences complement the aggregate results in \Cref{tab:terrain} by illustrating how LocoWM accommodates local support changes while retaining an upright payload.

\vspace{-5pt}
\vspace{-5pt}

\FloatBarrier
\subsubsection{Acceleration Compensation}
\label{subsubsec:app-acc-compensation}

\begin{figure}[htbp]
\centering
    \includegraphics[width=0.75\linewidth]{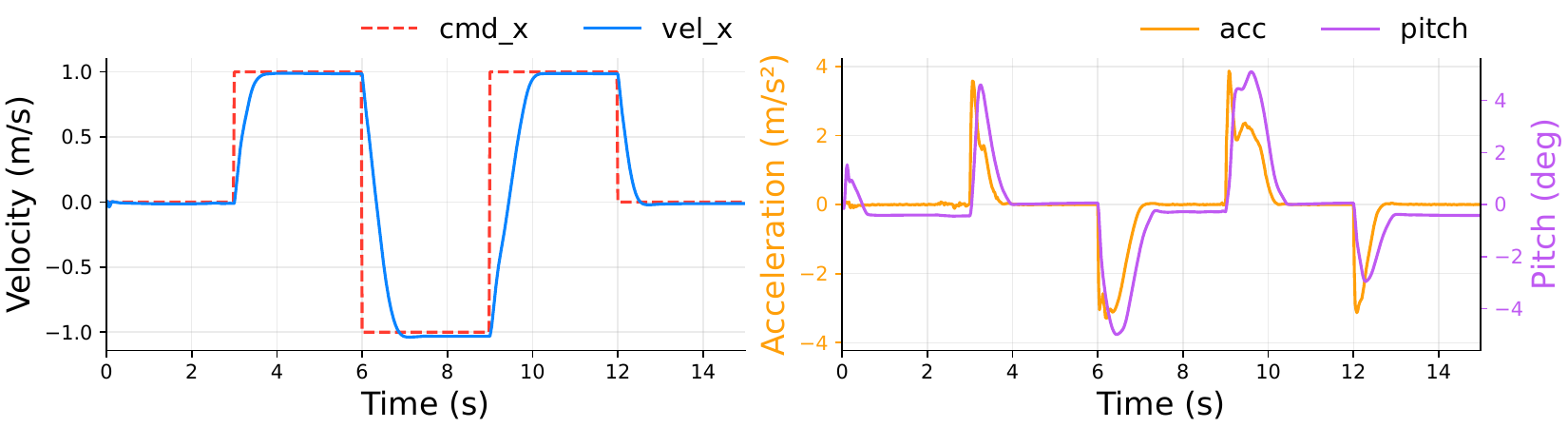}
    \caption{Simulated acceleration compensation under step velocity commands.}
\label{fig:app-sim-acc-curve-step}
\end{figure}

We examine how LocoWM adjusts platform pitch during acceleration and braking on flat ground using step and triangular-wave velocity commands.
In \Cref{fig:app-sim-acc-curve-step}, the commanded forward velocity follows the sequence $0 \rightarrow 1 \rightarrow -1 \rightarrow 1 \rightarrow 0$~m/s, and the measured velocity reaches each plateau after a brief transition.
Each command change produces an acceleration transient accompanied by a pitch response of the same sign.
Pitch returns close to zero as acceleration subsides, including during steady backward motion.
This response links platform tilt to acceleration while maintaining an approximately level posture at constant velocity.

\vspace{-5pt}
\vspace{-5pt}

\begin{figure}[htbp]
\centering
    \includegraphics[width=0.75\linewidth]{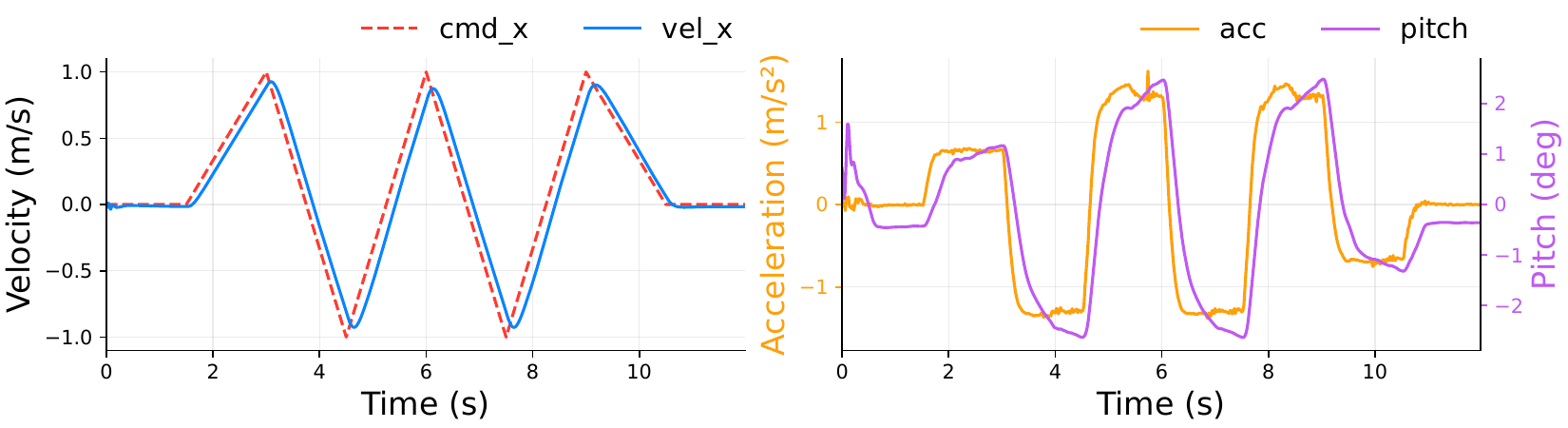}
    \caption{Simulated acceleration compensation under triangular-wave velocity commands.}
\label{fig:app-sim-acc-curve-triangle}
\end{figure}

\Cref{fig:app-sim-acc-curve-triangle} shows the response to triangular-wave commands spanning $-1$ to $1$~m/s.
The measured velocity follows the rising and falling ramps with a delay near their turning points.
During each ramp, sustained acceleration is accompanied by sustained platform tilt, with larger pitch magnitudes on the steeper ramps.
When the command slope changes sign, acceleration reverses and pitch subsequently transitions to the corresponding sign.

\vspace{-5pt}
\vspace{-5pt}

\begin{figure}[htbp]
\centering
    \includegraphics[width=0.85\linewidth]{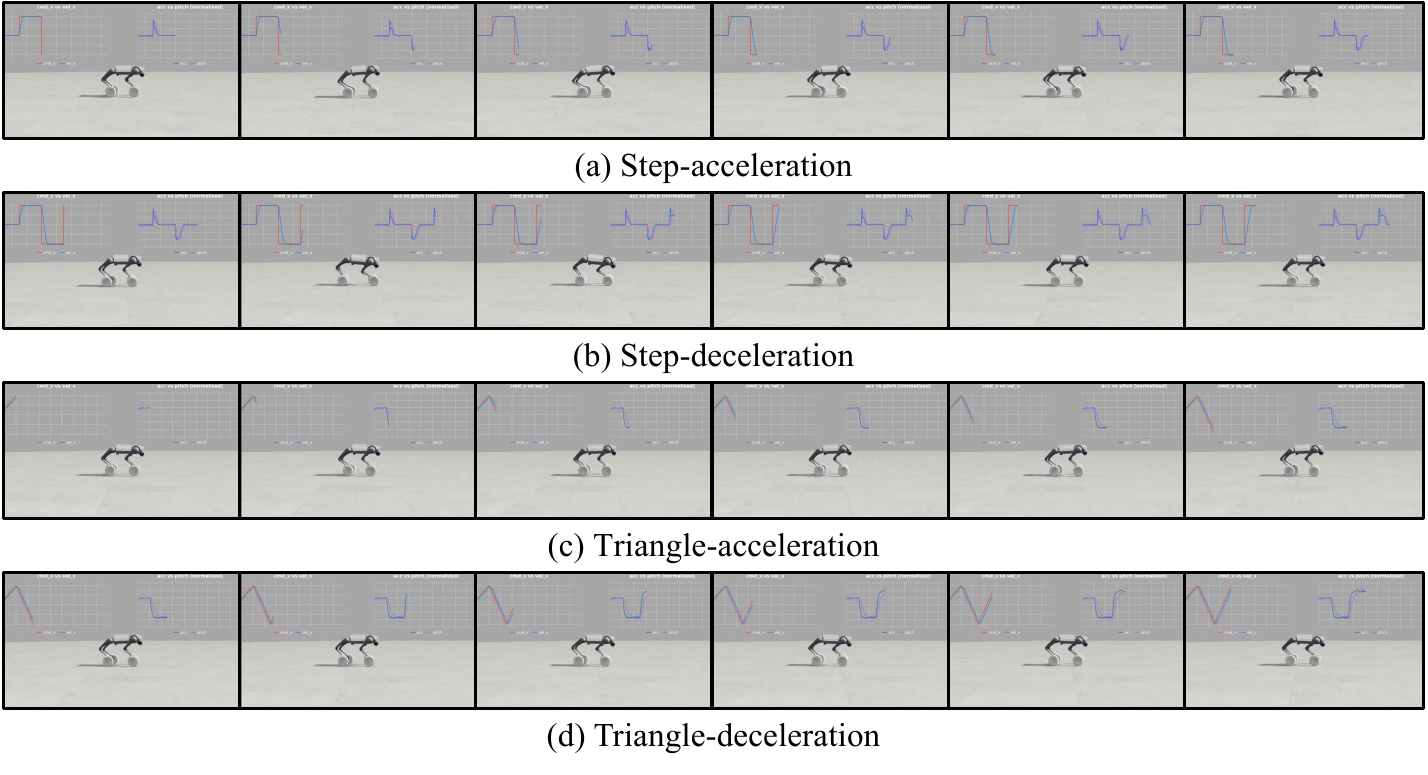}
    \caption{
        Simulated posture sequences during acceleration compensation under step commands in (a,b) and triangular-wave commands in (c,d).
        Insets show velocity tracking and normalized acceleration and pitch traces.
    }
\label{fig:app-sim-acc}
\end{figure}

The snapshots in \Cref{fig:app-sim-acc} show the leg configurations and torso poses associated with these responses.
Under step commands, the torso tilts during velocity transitions and returns toward level as velocity settles.
Under triangular-wave commands, the torso maintains a tilt during the ramps and changes its orientation as acceleration reverses.
Together, the curves and snapshots illustrate transient and sustained pitch adjustments for acceleration compensation.

\vspace{-5pt}
\vspace{-10pt}

\FloatBarrier
\subsubsection{Push Recovery}
\label{subsubsec:app-push}

\begin{figure}[htbp]
\centering
    \includegraphics[width=0.75\linewidth]{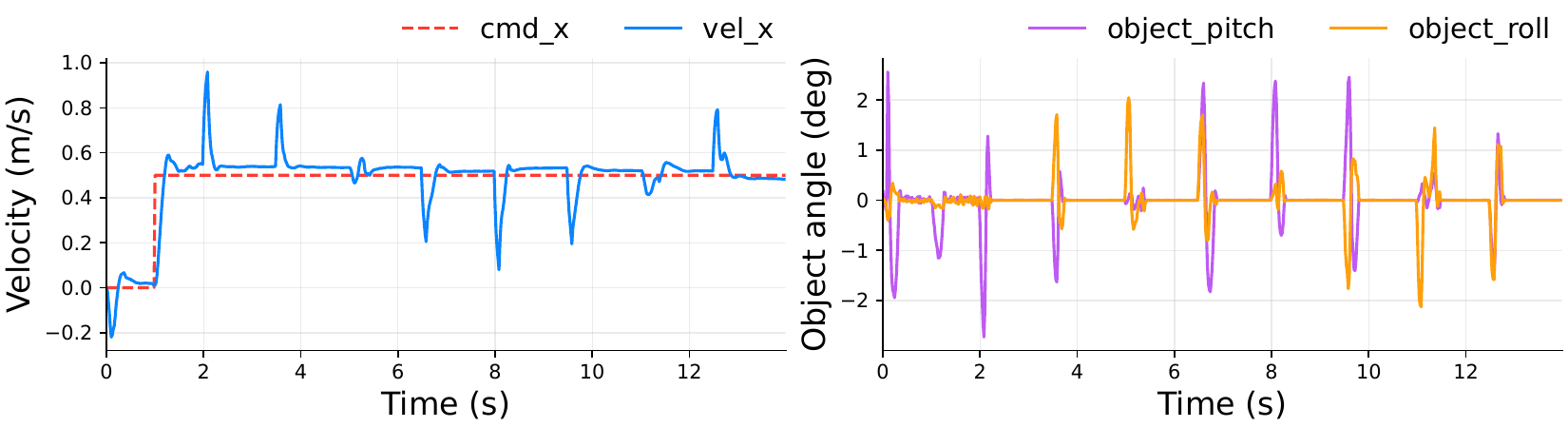}
    \caption{Simulated push recovery: forward velocity tracking (left) and payload pitch and roll (right).}
\label{fig:app-sim-push-curve}
\end{figure}

\Cref{fig:app-sim-push-curve} shows LocoWM's velocity and payload-orientation responses to repeated external pushes during forward locomotion.
After startup, the forward velocity command remains at $0.5$~m/s, while the disturbances produce transient increases and decreases in the measured velocity.
Velocity returns close to the command after each disturbance.
Payload pitch and roll undergo brief excursions of a few degrees and settle near zero between pushes, showing recovery of the payload's upright orientation alongside command tracking.

\vspace{-5pt}
\vspace{-5pt}

\begin{figure}[H]
\centering
    \includegraphics[width=0.75\linewidth]{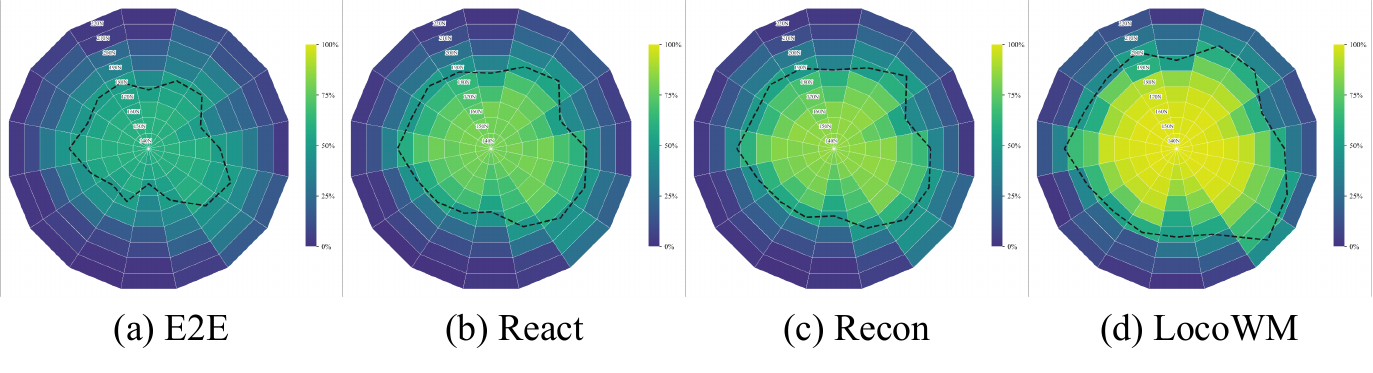}
    \caption{
        Payload-retention success-rate maps for (a) End-to-End (E2E), (b) React, (c) Recon, and (d) LocoWM.
        Polar angle denotes push direction, radius denotes force magnitude, and color denotes success rate.
        Dashed black contours mark a $50\%$ success rate.
    }
\label{fig:app-push-polar}
\end{figure}

\Cref{fig:app-push-polar} compares payload-retention success across push directions and force magnitudes for End-to-End, React, Recon, and LocoWM.
These maps complement the recovery sequences by showing how payload retention changes as disturbance strength and direction vary.

\vspace{-5pt}
\vspace{-10pt}

\begin{figure}[H]
\centering
    \includegraphics[width=0.85\linewidth]{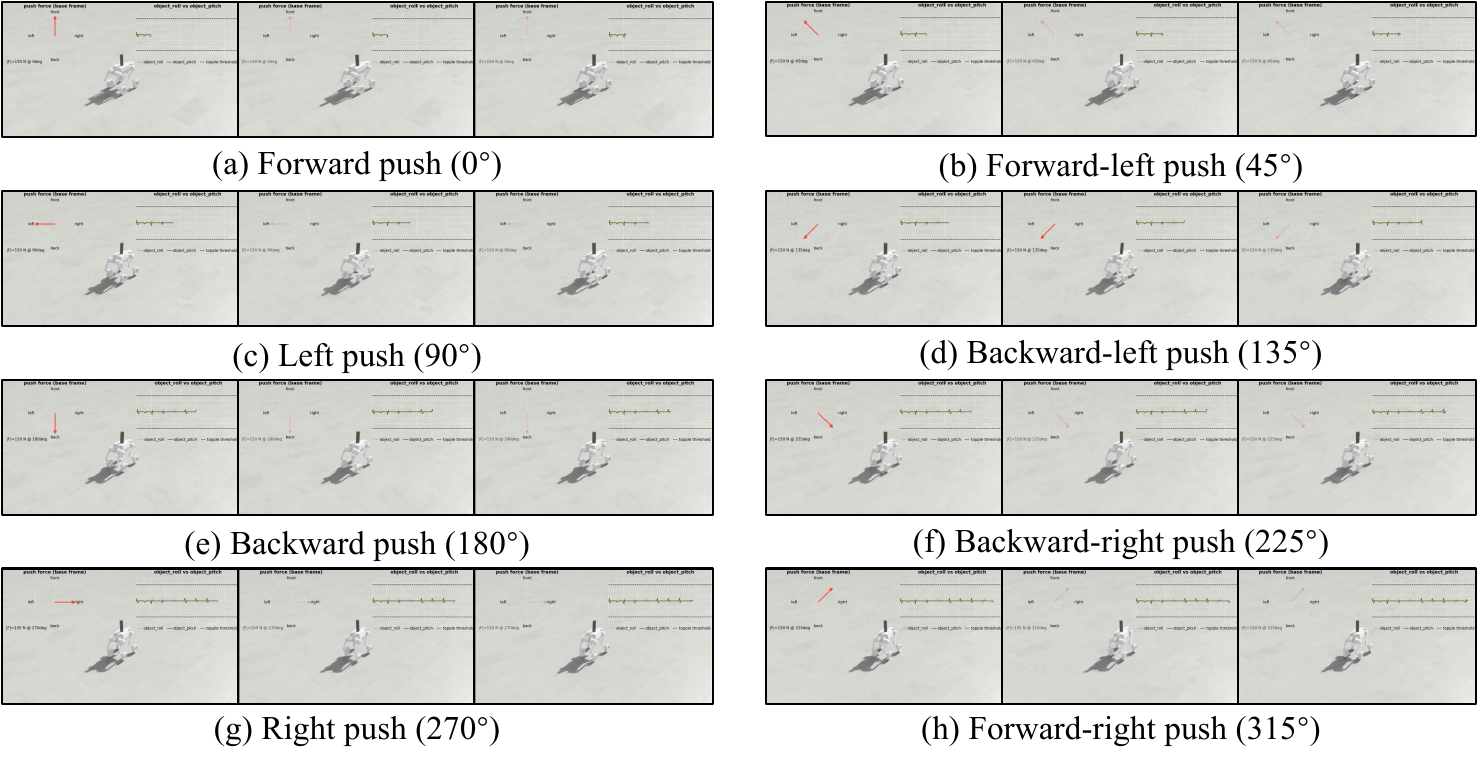}
    \caption{
        Simulated push recovery under $150$~N forces applied in eight body-frame directions.
        Frames progress from left to right through force application and recovery.
        Arrows indicate push direction, and inset traces show payload roll and pitch with the toppling thresholds.
    }
\label{fig:app-sim-push}
\end{figure}

The sequences in \Cref{fig:app-sim-push} show recovery from $150$~N pushes applied in eight directions, spaced at $45^\circ$ intervals in the body frame.
Each row follows force application and the subsequent recovery, with changes in the robot's posture and leg configuration as it responds to the disturbance.
The payload remains on the platform in all eight displayed sequences, and the inset orientation traces remain within the indicated toppling thresholds.

\subsection{Real-World Demonstrations}
\label{subsec:app-real}

\begin{figure}[H]
\centering
    \includegraphics[width=0.85\linewidth]{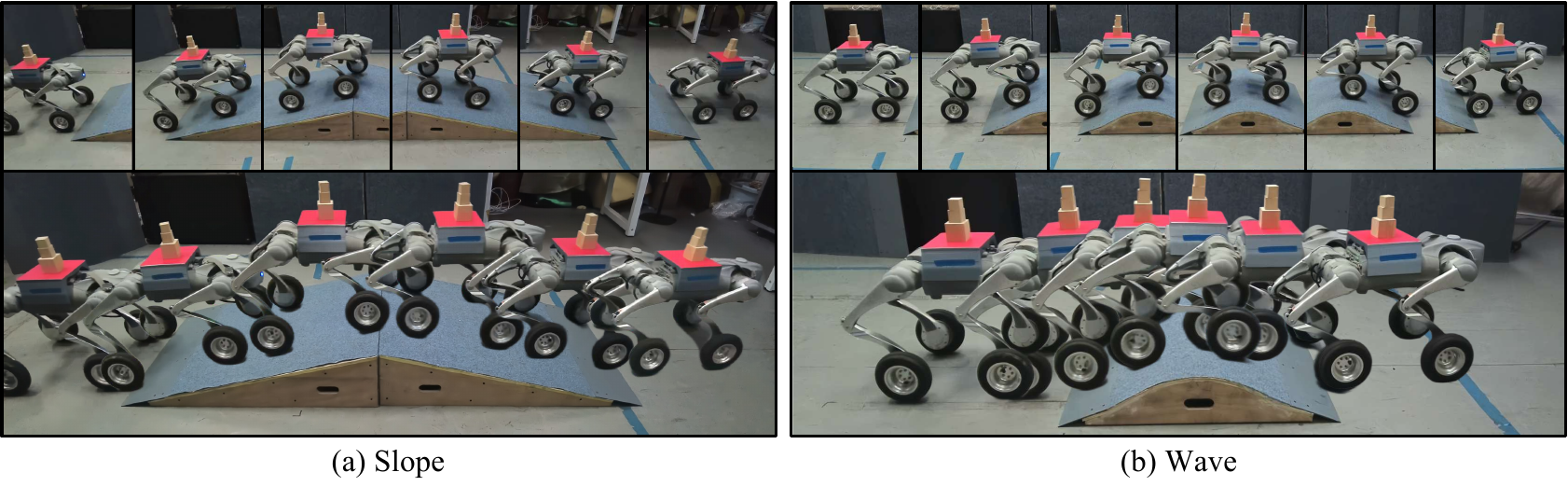}
    \caption{
        Real-world terrain leveling with stacked blocks on (a) a slope and (b) a wave-shaped obstacle.
        Each panel shows successive video frames above and robot cutouts overlaid on a common background below.
    }
\label{fig:app-real-leveling}
\end{figure}

\textbf{Terrain Leveling.}
Complementing the one-sided bridge and bump experiments in the main text, \Cref{fig:app-real-leveling} shows transport of unsecured stacked blocks over a slope and a wave-shaped obstacle.
LocoWM adjusts the leg configuration to changing support conditions, keeping the back-mounted platform approximately level in the displayed sequences.
Near the crest and obstacle exits, the front and rear wheels contact different terrain segments while the platform orientation remains distinct from the local ground inclination.
The successive video frames record the traversal, and the overlaid views show how wheel--leg configurations and payload orientation evolve along the path.
The block stack remains upright in both sequences, illustrating hardware leveling and payload retention under changing support heights.

\vspace{-5pt}
\vspace{-5pt}

\begin{figure}[H]
\centering
    \includegraphics[width=0.85\linewidth]{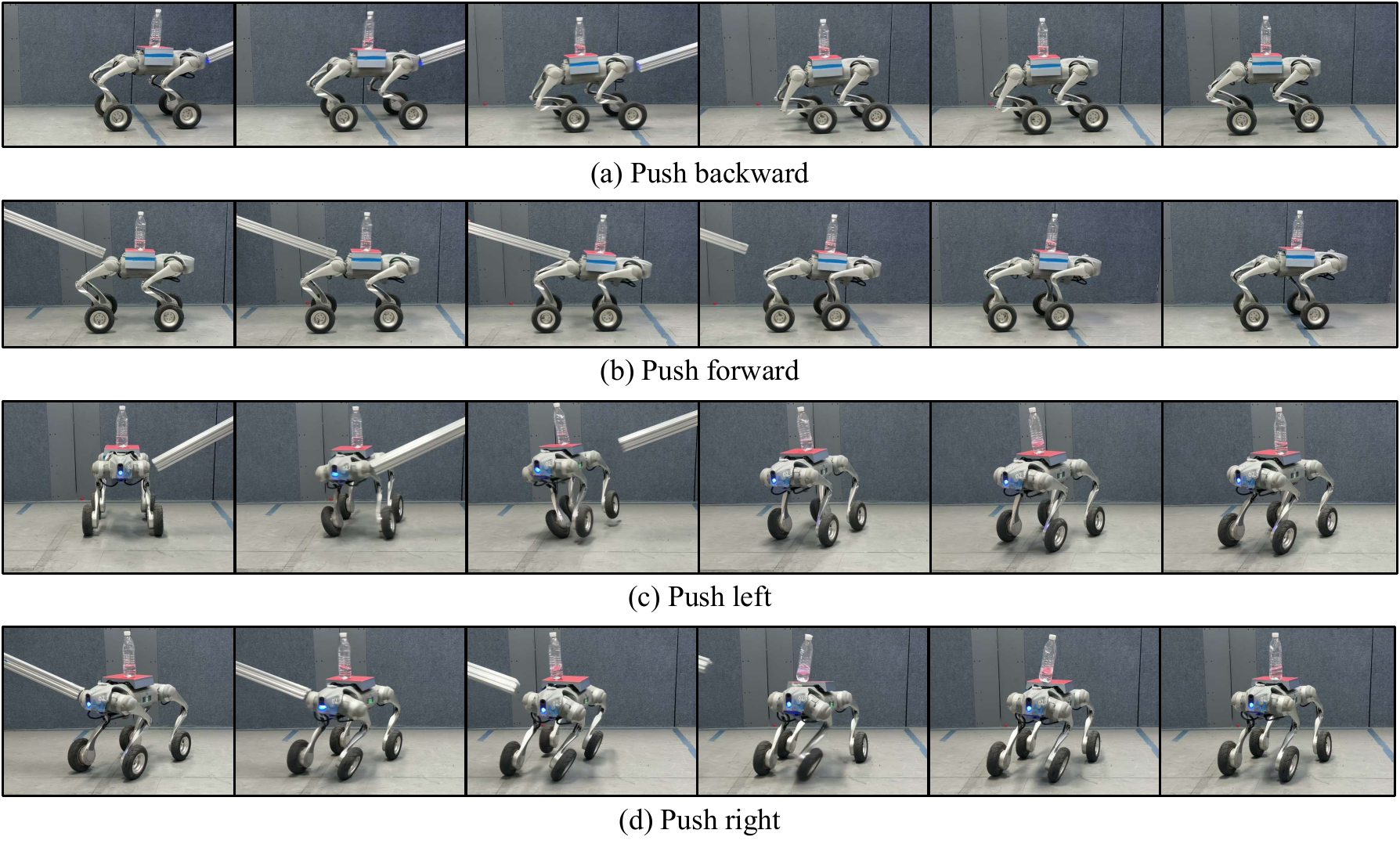}
    \caption{
        Real-world push recovery under (a) backward, (b) forward, (c) leftward, and (d) rightward pushes.
        Each row progresses from force application to recovery, with the unsecured bottle retained on the platform.
    }
\label{fig:app-real-push}
\end{figure}

\textbf{Push Recovery.}
\Cref{fig:app-real-push} shows responses to backward, forward, leftward, and rightward pushes applied to the torso with a pole while an unsecured bottle rests on the platform.
The side and frontal views capture the resulting pitch and roll disturbances, respectively.
LocoWM adjusts body posture and leg configuration during the disturbance and brings the platform back toward level after the push.
The bottle tilts transiently but remains on the platform without toppling in all four displayed sequences.

\vspace{-5pt}
\vspace{-5pt}

\begin{figure}[H]
\centering
    \includegraphics[width=0.85\linewidth]{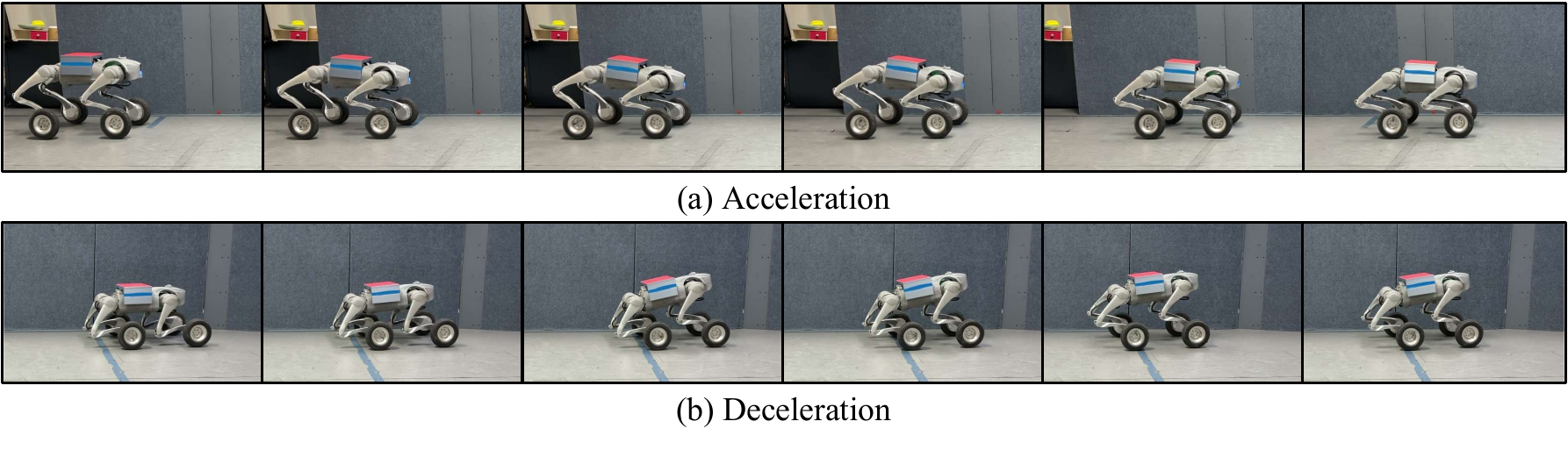}
    \caption{
        Real-world acceleration compensation during (a) acceleration and (b) deceleration on flat ground.
        Successive frames from left to right show changes in leg configuration and platform pitch.
    }
\label{fig:app-real-acc}
\end{figure}

\textbf{Acceleration Compensation.}
\Cref{fig:app-real-acc} supplements the hardware demonstrations in the main text with separate posture sequences during acceleration and deceleration on flat ground.
During acceleration, the torso adjusts its pitch as the motion changes and returns toward level in the later frames.
During deceleration, changes in the front and rear leg configurations accompany a corresponding adjustment in platform tilt.
Because the platform is rigidly attached to the torso, these orientation changes are achieved through the robot's own wheel--leg motion.
Both sequences illustrate active platform tilting during changes in longitudinal motion, consistent with the acceleration compensation behavior observed in simulation.

\FloatBarrier
\section{Humanoid Transport}
\label{sec:app-humanoid}

\subsection{Experimental Setup}
\label{subsec:app-humanoid-setup}

Following the SteadyTray task definition~\citep{huang2026steadytray}, we evaluate \emph{bimanual humanoid tray transport} in Isaac Lab with a Unitree G1 humanoid.
The G1 has 29 actuated joints, and both hands support a tray carrying an unsecured cylindrical payload.
The tray measures $25.4 \times 35.2$~cm, while the payload radius and height are randomized within $[2.1,4.5]$~cm and $[7.5,20]$~cm, respectively.

\textbf{Task.}
We evaluate three scenarios: \emph{Command Track}, \emph{Push Robot}, and \emph{Push Object}.
The velocity command is zero during the first second, switches to a sampled target during $1$--$10$~s, and switches once more at $10$~s.
For the two push scenarios, a constant world-frame $XY$ force is applied for $0.1$~s at $5$, $10$, and $15$~s.
The nominal per-axis velocity-change bound is $0.5$~m/s for Push Robot and $0.3$~m/s for Push Object, while Command Track has no external push.

\textbf{Baselines.}
The evaluation compares three executable controller paths: Base-Policy, ReST-RL, and LocoWM.
Base-Policy denotes the deterministic output of the frozen whole-body locomotion policy.
ReST-RL uses the deployable residual branch from the public checkpoint.
The following table therefore reports a reconstruction, rather than an exact reproduction of every method row in the original Table~II in \citep{huang2026steadytray}.

\textbf{Metrics.}
TrackLinErr is the $L_2$ error between commanded and measured horizontal velocity in the torso frame.
Grav-XY is the norm of the $XY$ components of unit gravity expressed in the object frame.
A trial is successful only if the object remains above $0.7$~m and its tilt remains below $0.7$~rad throughout the full horizon.

\subsection{Comparison Results}
\label{subsec:app-humanoid-results}

\begin{table}[t]
\centering
\small
\setlength{\tabcolsep}{3.5pt}
\begin{tabular}{@{}llccccc@{}}
\toprule
\multirow{2}{*}{Task} &
\multirow{2}{*}{Method} &
\multicolumn{2}{c}{TrackLinErr (m/s) $\downarrow$} &
\multicolumn{2}{c}{Grav-XY $\downarrow$} &
\multirow{2}{*}{Success Rate (\%) $\uparrow$} \\
\cmidrule(lr){3-4}
\cmidrule(lr){5-6}
& & mean & std & mean & std & \\
\midrule
\multirow{3}{*}{Command Track}
& Base-Policy 
& 0.128 & 0.096 & 0.498 & 0.431 & 52.7 \\
& ReST-RL
& 0.123 & 0.108 & 0.083 & 0.212 & 91.0 \\
& LocoWM
& \textbf{0.120} & 0.102 & \textbf{0.063} & 0.165 & \textbf{94.1} \\
\midrule
\multirow{3}{*}{Push Robot}
& Base-Policy 
& 0.161 & 0.162 & 0.751 & 0.388 & 12.5 \\
& ReST-RL
& \textbf{0.150} & 0.148 & \textbf{0.128} & 0.281 & 80.9 \\
& LocoWM
& 0.155 & 0.165 & 0.129 & 0.280 & \textbf{82.8} \\
\midrule
\multirow{3}{*}{Push Object}
& Base-Policy 
& 0.130 & 0.105 & 0.567 & 0.435 & 44.1 \\
& ReST-RL
& 0.123 & 0.107 & 0.082 & 0.209 & 90.2 \\
& LocoWM
& \textbf{0.122} & 0.112 & \textbf{0.064} & 0.163 & \textbf{92.6} \\
\bottomrule
\end{tabular}\caption{
  Simulation results for humanoid tray transport.
  Bold entries mark the best mean error or success rate within each task.
}
\label{tab:app-humanoid-results}
\end{table}

Table~\ref{tab:app-humanoid-results} shows that both ReST-RL and LocoWM achieve higher Success Rate than Base-Policy across all three tasks.
LocoWM achieves the highest Success Rate of $94.1\%$, $82.8\%$, and $92.6\%$ in Command Track, Push Robot, and Push Object, exceeding ReST-RL by $3.1$, $1.9$, and $2.4$ percentage points, respectively.
LocoWM also achieves the lowest mean TrackLinErr and Grav-XY in both Command Track and Push Object.
In Push Robot, ReST-RL attains slightly lower mean TrackLinErr and Grav-XY, whereas LocoWM achieves a higher Success Rate, increasing it from $12.5\%$ with Base-Policy to $82.8\%$.
These results show that LocoWM achieves higher transport success across the evaluated scenarios, although it does not attain the lowest mean tracking error and payload tilt metric in every scenario.

\subsection{Visualization}
\label{subsec:app-humanoid-visualization}

\begin{figure}[H]
\centering
    \includegraphics[width=0.75\linewidth]{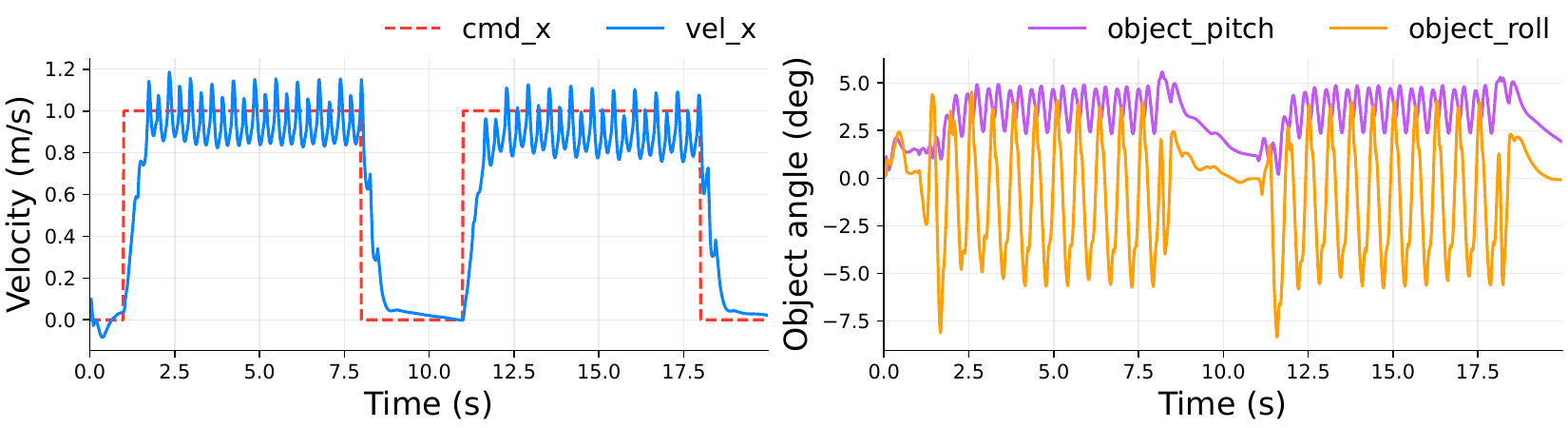}
\caption{Simulated humanoid transport during starting, walking, and stopping.}
\label{fig:app-sim-go1-curve}
\end{figure}

\Cref{fig:app-sim-go1-curve} shows two walking intervals separated by a stop, with the forward velocity command switching between $0$ and $1$~m/s.
The measured velocity rises after each start command, oscillates during walking, and returns close to zero after each stop command.
Payload pitch and roll also exhibit periodic oscillations during walking.
These oscillations subside after stopping as the payload settles.

\begin{figure}[H]
\centering
    \includegraphics[width=0.85\linewidth]{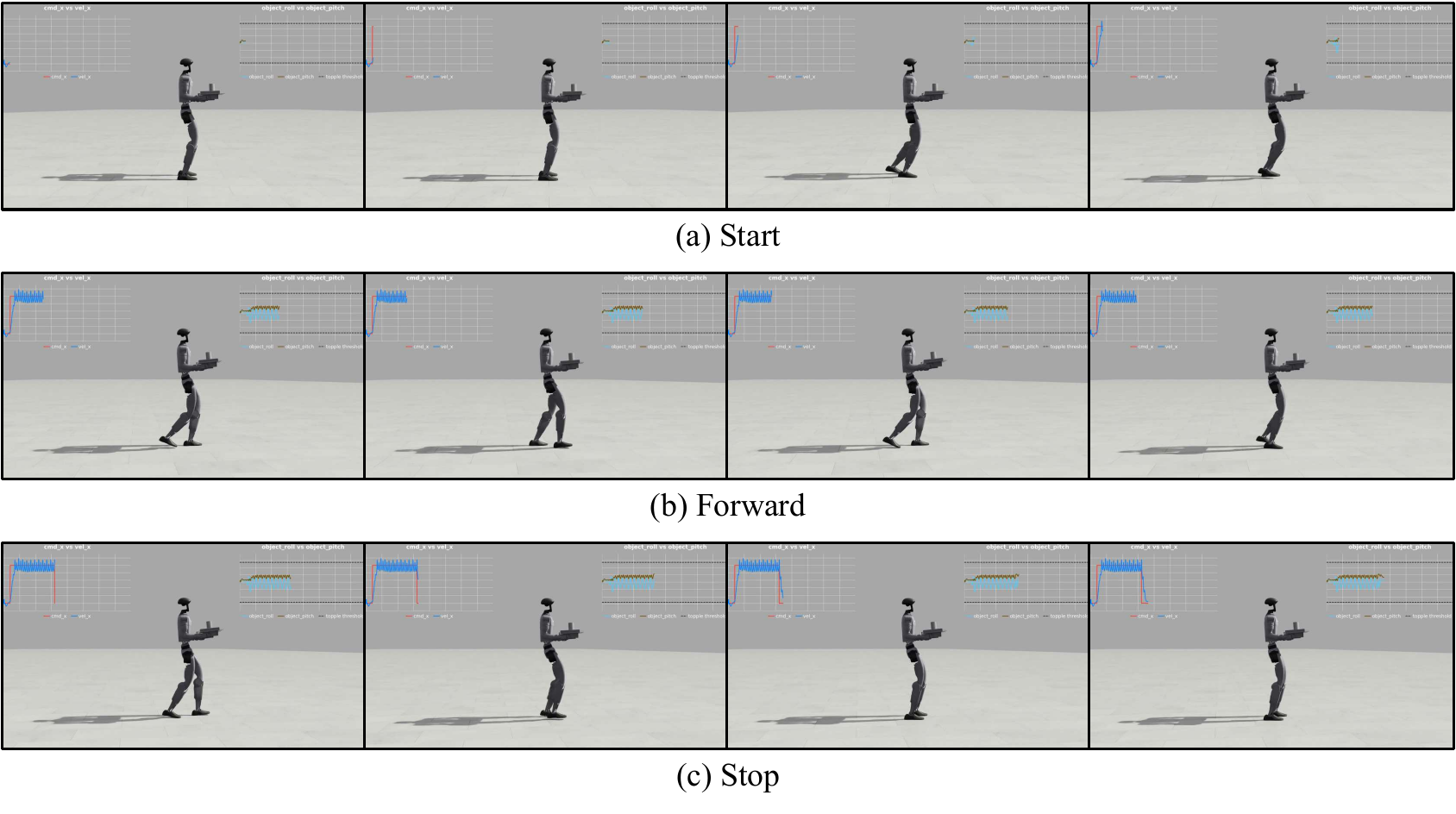}
\caption{
Simulated humanoid tray transport during (a) starting, (b) forward walking, and (c) stopping.
Frames progress from left to right, with insets showing velocity tracking and payload orientation.
}
\label{fig:app-sim-go1}
\end{figure}

The snapshots in \Cref{fig:app-sim-go1} show the corresponding transitions from standing to stepping, continued forward walking, and a return to standing.
The robot alternates leg support while holding the tray in front of its torso, exposing the payload to changing body motion and contact conditions.
The payload remains on the tray without toppling throughout the displayed sequences.
This example complements the Go2-W experiments by applying preactive residual control to end-effector stabilization during bipedal locomotion.

\end{document}